\documentclass{article} 
\usepackage{arxiv,times}

\usepackage{amsmath,amsfonts,bm}

\def\eqref#1{equation~\ref{#1}}

\def\1{\bm{1}}

\DeclareMathAlphabet{\mathsfit}{\encodingdefault}{\sfdefault}{m}{sl}
\SetMathAlphabet{\mathsfit}{bold}{\encodingdefault}{\sfdefault}{bx}{n}

\usepackage{hyperref}
\usepackage{url}
\usepackage{graphicx}
\usepackage{booktabs} 
\usepackage{pifont}
\usepackage{cleveref}
\usepackage{comment}
\usepackage{multirow}
\usepackage{subcaption}
\usepackage{eccvabbrv}
\usepackage{threeparttable}
\usepackage{float}
\title{HyperFlow: Class-Incremental Semantic Change Detection via Hyperbolic Flow}

\author{Manon Béchaz\textsuperscript{1}\footnote{Correspondence: manon.bechaz@epfl.ch}, Léo Buecher\textsuperscript{2,3}, Emanuele Dalsasso\textsuperscript{4}, Ciprian Tomoiaga\textsuperscript{5}, Patrick Jayet\textsuperscript{5}, \\ \textbf{Marcin Detyniecki\textsuperscript{5}, Laetitia Chapel\textsuperscript{3,6}, Lucas Drumetz\textsuperscript{7}, Nicolas Courty\textsuperscript{2,3}, Devis Tuia\textsuperscript{1}}  \\
\textsuperscript{1}EPFL, Switzerland;  \textsuperscript{2}Université de Bretagne Sud, France; \textsuperscript{3}IRISA, France \textsuperscript{4}INRIA, France;\\ \textsuperscript{5}AXA GDAI  Research, Switzerland;
\textsuperscript{6}Institut Agro Rennes-Angers, France \textsuperscript{7}Lab-STICC,\\ IMT Atlantique, France. 
Correspondence : \texttt{manon.bechaz@epfl.ch} \\}
\preprintcopy
\begin{document}

\maketitle

\begin{abstract}

Semantic change detection in remote sensing aims to identify and characterize land-cover transitions  observed in pairs of satellite or aerial images. Existing methods assume a fixed set of change categories, limiting their applicability in real-world scenarios where changes of interest may vary, for example across geographies or over time. In this work, we tackle the problem of class-incremental semantic change detection, where a model progressively learns to detect new types of semantic change. To do so, we exploit the hierarchical structure of land-cover semantics, where newly introduced categories correspond to fine-grained refinements of existing ones, and represent said semantic categories in hyperbolic space, which is particularly well suited to represent such hierarchical relationships. To capture semantic changes, we introduce HyperFlow, a method relying on Busemann prototypes, modeling change as an anomaly in a learned semantic flow within the hyperbolic embedding space induced by segmentation. HyperFlow couples change detection to class-incremental semantic segmentation, enabling class-incremental semantic change detection. 
We also present HiChange, a new large-scale, high resolution remote sensing dataset with detailed, fine-grained hierarchical labels. Experimental results demonstrate that HyperFlow achieves both strong change detection and semantic segmentation performance across incremental steps. In particular, we show that progressively refining semantic categories, rather than considering them all directly, yields improved change detection, highlighting the benefits of incremental learning for semantic change detection.
\end{abstract}

\section{Introduction}
\label{sec:intro}
Semantic change detection (SCD) aims at identifying and characterizing land cover transitions over time by comparing pairs of co-registered satellite or aerial images. Despite being a well established research field \citep{jiang2022survey} with applications in disaster response \citep{zheng2021building,ahn2025generalizable}, urban planning \citep{wang2024rsbuilding,zhao2025df2net,zheng2021building} or land cover analysis \citep{li2016superresolution,lv2023spatial,daudt2019multitask}, current approaches remain static and assume a fixed set of change categories. This assumption is misaligned with real-world scenarios, where changing environmental and societal priorities imply evolving monitoring objectives \citep{tasar2019incremental,weng2024mdinet,zhang2023lim}. Adapting models to new types of changes typically requires costly retraining and annotation, making current approaches unsuitable for long-term deployment.
\begin{figure}[t]
    \centering
    \includegraphics[width=\linewidth]{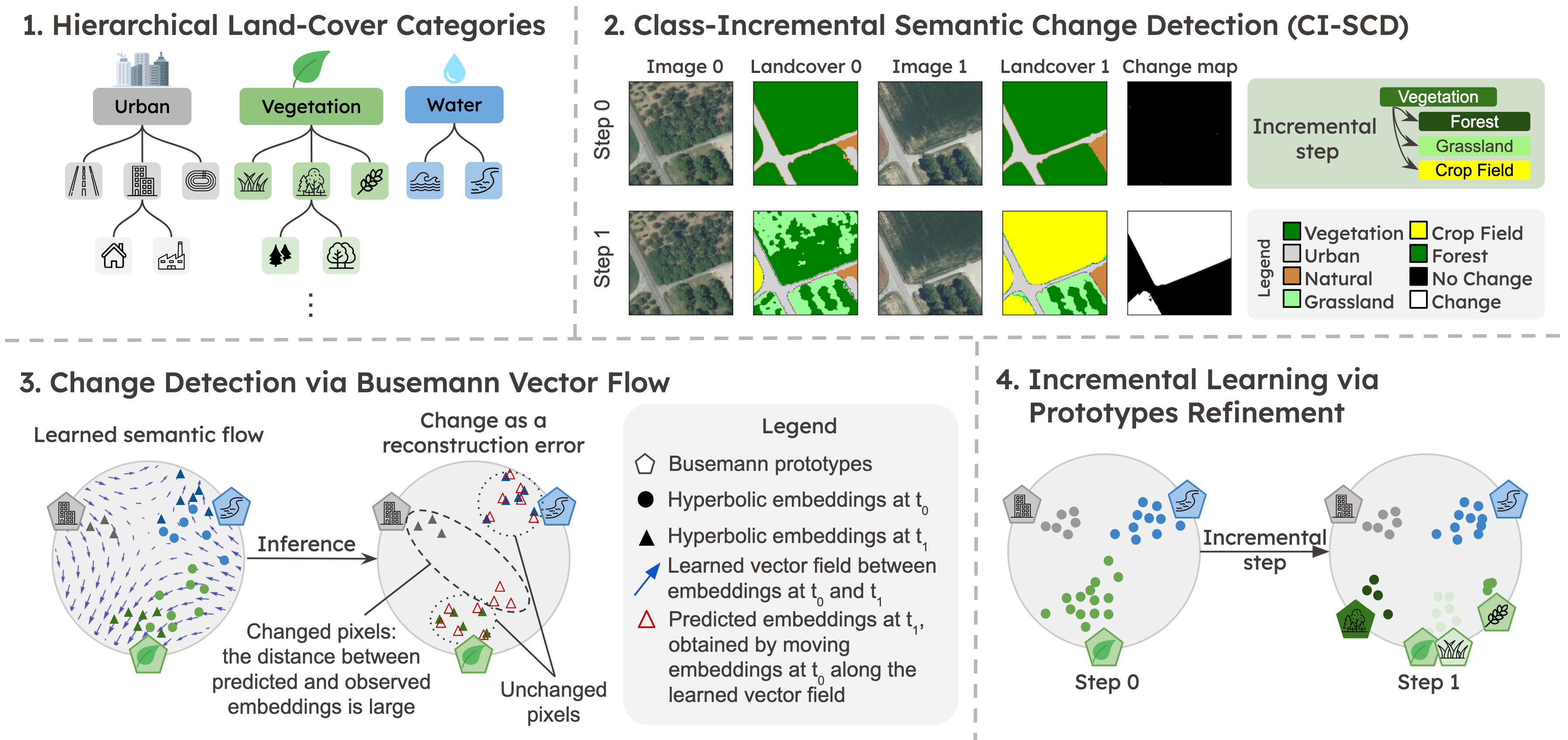}
    \caption{Overview of CI-SCD via Hyperbolic Flow. (1) Land-cover semantics are naturally organized in hierarchies. (2)
    During CI-SCD, the model leverages this taxonomy to progressively refine semantic categories from coarse to fine-grained classes. Change maps evolve to account for the progressively refined segmentation. (3) Hyperbolic semantic flow: a vector field transforming the embeddings of the first image into the embeddings of the second image is learned. The hyperbolic distance between the positions of the predicted embeddings and the observed positions indicates change. (4) Prototype-based incremental learning: we refine a coarse class by splitting it into leaves prototype nodes. The vector field in (3) is then retrained to adapt to the new semantic classes.}
    \label{fig:introduction}
\end{figure}
A key observation is that land-cover categories are naturally structured in hierarchies, from broad categories like \textit{urban} or \textit{vegetation} to fine-grained ones such as \textit{residential building}, \textit{crop field}, or \textit{forest}. In the context of land use monitoring, new types of changes are rarely entirely new categories; rather, they usually correspond to refining coarse classes into more specific ones, as illustrated in \cref{fig:introduction}, panel (1). Leveraging this hierarchy is therefore essential to enable adaptable SCD methods and calls for change detection (CD) models that \textit{(i)} explicitly represent hierarchical semantics and \textit{(ii)} adapt to different levels of this hierarchy over time.

To tackle these challenges, we consider the task of class-incremental semantic change detection (CI-SCD), in which a model continuously expands the types of semantic changes it detects. First, we model land-cover semantics in hyperbolic space, which is particularly well suited to represent hierarchical structures \citep{nickelPoincareEmbeddingsLearning2017,ganeaHyperbolicEntailmentCones2018,ganeaHyperbolicNeuralNetworks2018}. 
Second, we leverage advances in class-incremental semantic segmentation (CISS) to advance CI-SCD. 
We introduce a dynamic hyperbolic formulation of CI-SCD, where semantic categories are obtained from hierarchical embeddings in hyperbolic space, while change is modeled as an anomaly in a learned semantic flow within this structured space.
This approach couples SCD to CISS, enabling CI-SCD, and ensures that the detected changes are adapted to the updated semantic classes (see \cref{fig:introduction}, panel (2)).

Specifically, we represent pixels at different time steps as points in a shared hyperbolic embedding space induced by the segmentation. We then model change as an anomaly in a learned semantic flow in this space, i.e., a displacement between embeddings at times $t_0$ and $t_1$: when a pixel changes (e.g., from \textit{vegetation} to \textit{building}), its embedding moves between regions of the manifold, producing a displacement in embedding space between $t_0$ and $t_1$, as illustrated in \cref{fig:introduction}, panel (3). By learning the expected flow accounting for small domain shifts naturally occurring between images (\eg due to seasonal or illumination changes), the remaining deviations indicate the unexpected changes to be detected. To learn this semantic flow in hyperbolic space, we introduce HyperFlow, a method based on the Busemann function. 
Building on prior work in hyperbolic segmentation \citep{ghadimiatigh_hyperbolic_2022,berg2024horospherical}, CISS \citep{hindel_taxonomy-aware_2024} and Riemannian vector fields \citep{katsmanRiemannianResidualNeural2023}, we show that Busemann prototypes are the key to link this approach to CISS within a single hyperbolic method. 
In particular, incremental learning is handled directly through those prototypes, as depicted in \cref{fig:introduction}, panel (4): when a coarse category is refined, its prototype is split to initialize fine-grained subclasses, preserving geometric proximity to the parent while enabling specialization. 
The semantic flow is then retrained in this updated space, thus becoming specific to the newly introduced subclasses and allowing the detection of new semantic transitions.

In summary, this work makes three main contributions. First, we propose HyperFlow, a method that represents semantic categories using Busemann prototypes and models change via a semantic flow in a shared embedding space, coupling segmentation and CD in a geometrically grounded manner. 
Second, we show how HyperFlow can be leveraged to address the task of CI-SCD by extending CISS methods. In this formulation, incremental refinement is achieved by splitting parent prototypes to initialize newly introduced subclasses, preserving hierarchical structure while ensuring that the semantic flow, retrained to account for this new embedding space, adapts consistently as the taxonomy evolves. Third, as current benchmarks in SCD are limited to a small number of generic semantic categories \citep{zhang2025foba} lacking hierarchical structure, we present HiChange, a large-scale dataset with a rich hierarchical label structure designed to enable CI-SCD models to detect fine grained changes of increasing semantic complexity.

\section{Related Work}
\label{sec:related_work}
\subsection{Semantic Change Detection in Remote Sensing}

CD enables the identification of differences between remotely sensed images of the same location acquired at different times. 
Initially based on hand-crafted features \citep{malila1980change,volpi2012unsupervised}, recent methods leverage deep learning, 
typically with Siamese designs \citep{daudt2018fully,jiang2022survey}. 
However, such approaches rely on the quality of ground-truth change maps, which are difficult to obtain and often noisy \citep{daudt2019multitask}. To overcome the need for accurate change labels, unsupervised CD has been widely studied \citep{malila1980change,nielsen1997multivariate, volpi2012unsupervised}, and is currently framed as anomaly detection, linking changes to high reconstruction errors \citep{touati2020anomaly,andresini2022leveraging,bechaz20262player,ruuvzivcka2022ravaen,noh2022unsupervised,liu2021unsupervised,luppino2021deep}.
While most methods focus on binary change detection (BCD), real world scenarios require specifying the semantics of change. 
SCD therefore extends BCD by assigning semantic labels to changes  \citep{daudt2019multitask}. Early SCD methods used post-segmentation comparisons 
\citep{huang2010sampling,hu2013seasonal,wu2017post} and suffered from error accumulation and lack temporal modeling \citep{volpi2013supervised}. 
Deep learning SCD methods have since become dominant and adopt end-to-end architectures that couple semantic segmentation and change detection \citep{daudt2019multitask,zhu2022land,ding2024joint}. 
Despite this progress, existing SCD methods remain rigid: they assume fixed categories, and treat them as independent labels, ignoring the hierarchical structure that naturally underlies land-cover semantics. These assumptions are unrealistic in real-world settings, where categories are progressively refined and semantically nested, and motivate our study of CI-SCD.



\subsection{Class-Incremental Semantic Segmentation}

CISS addresses the problem of extending a segmentation model to new categories without forgetting previously learned ones. Existing CISS methods are replay-based, expansion-based, and distillation-based. Replay-based methods store past examples~\citep{cha2021ssul},  or encode them using prototypes~\citep{chen2023saving,zhu2025adaptive} and generative models~\citep{maracani2021recall}. Expansion-based approaches expand the model architecture to allocate components for new classes, e.g. via parallel branches or class-specific modules \citep{zhang2022representation,xiao2023endpoints},  or by preserving prior knowledge by freezing parameters associated with previously learned classes \citep{srivastava2019adaptive,cha2021ssul}. Distillation-based methods constrain the updated model to preserve the previous responses on old classes, e.g. by distilling logits \citep{cermelli2020modeling,michieli2019incremental,baek2022decomposed}, intermediate features \citep{douillard2021plop,michieli2019incremental} or attention relations \citep{qiu2023sats}. 
Moreover, recent work showed that explicitly modeling semantic hierarchies, particularly using hyperbolic embedding spaces, significantly improves class-incremental segmentation via structured refinement \citep{hindel_taxonomy-aware_2024,lin2022continual}. In this work, we build upon these advances in CISS 
to address the specific problem of CI-SCD, where the taxonomy of the semantic classes is progressively refined on a corpus of images that remains fixed across incremental steps.

\subsection{Hyperbolic Spaces}

Hyperbolic geometry is particularly adapted to model hierarchical labels \citep{nickelPoincareEmbeddingsLearning2017,ganeaHyperbolicEntailmentCones2018,ganeaHyperbolicNeuralNetworks2018}, with applications in classification \citep{khrulkov_hyperbolic_2020}, object detection \citep{liu_hyperbolic_2020,lang_hyperbolic_2022}, and semantic segmentation~\citep{weber2024flattening}. \citet{ghadimiatigh_hyperbolic_2022} showed that hyperbolic multinomial logistic regression improves performance and hierarchical consistency, particularly in low dimension, while \cite{hindel_taxonomy-aware_2024} showed the importance of leveraging taxonomies. 
In this context, Busemann functions have shown promise \citep{chamiHoroPCAHyperbolicDimensionality2021, fanHorosphericalDecisionBoundaries2023, ghadimi2021hyperbolic, berg2024horospherical, bergMultiPrototypeHyperbolicLearning2025}, extracting geometrically meaningful features that quantify dissimilarity to ideal prototypes located at infinity. 
More recently, hyperbolic spaces have been explored for class-incremental settings, both in image classification \citep{cui_rethinking_2024} and taxonomy-aware segmentation \citep{hindel_taxonomy-aware_2024}, showing that hyperbolic embeddings can mitigate catastrophic forgetting by preserving hierarchical relations between classes. In our setting, where incremental refinement corresponds to splitting coarse land-cover categories into finer-grained ones, such hierarchical modeling is particularly well suited to maintain semantic consistency. We therefore adopt Busemann prototypes \citep{berg2024horospherical, bergMultiPrototypeHyperbolicLearning2025} not only to structure hierarchical segmentation, but also to define the geometric basis of our semantic flow formulation for CI-SCD.

\subsection{Riemannian Vector Fields}


A natural way to model displacements in hyperbolic space is through vector fields
and their flow. Parameterizable vector fields on manifolds for deep learning tasks were first considered for extending neural Ordinary Differential Equations to the Riemannian setting \citep{rezendeNormalizingFlowsTori2020, falorsiNeuralOrdinaryDifferential2020, mathieuRiemannianContinuousNormalizing2020, louNeuralManifoldOrdinary2020, katsmanEquivariantManifoldFlows2021}. 
While most approaches rely on extrinsic, Euclidean, formulations, ignoring the geometry of the manifold, \citet{katsmanRiemannianResidualNeural2023} designed intrinsic vector fields in hyperbolic space for Riemannian ResNets, relying on the Busemann functions. Other designs were specifically investigated for images \citep{vanspenglerPoincareResNet2023}. In this work, we adopt the approach of \citet{katsmanRiemannianResidualNeural2023}: we construct a vector field induced by the Busemann functions, in order to model natural shifts of the hyperbolic embeddings between the temporal images.

\section{Methodology}
\label{sec:methodology}

\subsection{Class-Incremental Semantic Change Detection} 

Let $I_0,I_1\in\mathbb{R}^{H\times W\times C_\mathrm{in}}$ denote a co-registered pair of remote sensing (RS) images acquired at two times $t_0$ and $t_1$. SCD aims at characterizing, for each changed pixel, the type of change. 
This is achieved by predicting a binary change map $C\in\{0,1\}^{H\times W}$, indicating which pixels changed, together with semantic labels $Y_0,Y_1\in\mathcal{Y}^{H\times W}$ detailing the change type \citep{daudt2019multitask}.

Traditional methods assume that the semantic label space $\mathcal{Y}$ is fixed and fully specified during training. 
Instead, we consider a class-incremental scenario where categories are progressively refined over a sequence of incremental steps $s\in\{0,...,n\}$. Let $T_{K_s}^s$ be the task linked to the $s$-th incremental step associated with a task-specific label set $\mathcal{Y}_s$, of size
$K_s=|\mathcal{Y}_s|$.
This label space forms a taxonomic tree $\mathcal{H}$  whose internal nodes represent coarse concepts (\eg, vegetation, built-up), and  leaves represent fine-grained ones (\eg, coniferous forest, lake), see \cref{fig:introduction}, panel (1). The same input images are considered for all incremental steps, but for each task the labels are drawn from the task-specific set $\mathcal{Y}_s\subseteq\mathcal{H}$. Note that 
new labels in $\mathcal{Y}_s$ arise from refinements of existing categories and are appended to the existing ones. For each task $T_{K_{s-1}}^{s-1},\:s>0$, if a coarse class $k\in\mathcal{Y}_{s-1}$ is refined into children $\mathrm{child}(k)=\{k_1,…,k_m\}$, then for the task $T_{K_s}^s$ the label set is  \begin{equation}
    \mathcal{Y}_s =\mathcal{Y}_{s-1}\setminus k \cup \mathrm{child}(k).
\end{equation}
Incremental steps affect both the semantic segmentation and the CD: as the semantic label space is refined, the change maps are correspondingly updated to reflect transitions between categories in $\mathcal{Y}_s$ only, as depicted in \cref{fig:introduction}, panel (2). 

\subsection{Hyperbolic Semantic Segmentation}\label{sec:method_hyperbolic}

To tackle semantic segmentation in hyperbolic space, we adopt the horospherical classifier from~\cite{berg2024horospherical}. It is based on Busemann prototypes, that play a central role in the change detection mechanism introduced later in \cref{sec:method_busemann_flow}.
We use a segmentation backbone where only the last network layer is mapped to hyperbolic space \citep{hindel_taxonomy-aware_2024,ghadimiatigh_hyperbolic_2022,khrulkov_hyperbolic_2020}.
Following \cite{hindel_taxonomy-aware_2024}, we use the DeepLabV3 model with a ResNet-101 backbone pre-trained on ImageNet, and consider the $d$-dimensional Poincaré ball model of hyperbolic space 
defined as $\mathbb{B}_d = \{z\in\mathbb{R}^d\:|\:\|z\|^2<1\}$. Unless stated otherwise, $d=2$, which already provides competitive performance compared to baselines. Experiments in higher dimensions and with other backbones are in the Supplementary Materials \ref{supp:influence_backbone} and \ref{sec:supp_dim}.

Let $e$ denote the Euclidean embedding vector produced by the segmentation backbone for pixel $x$. These embeddings are mapped onto the Poincaré ball via the exponential map at the origin:\begin{equation}
    z=\mathrm{exp}_0(e)=\mathrm{tanh}\big(\|e\|_2\big)\cdot e/\|e\|_2 \in\mathbb{B}_d.
\end{equation}
Each semantic class $k\in\mathcal{Y}_s$ is parameterized by a Busemann prototype $p_k\in\mathbb{S}^{d-1}$ located on the boundary of the Poincaré ball together with a scalar bias $a_k\in\mathbb{R}$,  both learned during training. They define a horosphere, which can be interpreted as the hyperbolic analog of a Euclidean hyperplane:\begin{equation}
    H_k=\left\{z\in\mathbb{B}_d \:|\: -B_{p_k}(z)+a_k=0\right\},
\end{equation}where $B_{p_k}$ is the Busemann function \citep{chamiHoroPCAHyperbolicDimensionality2021,bridsonMetricSpacesNonPositive1999}:
\begin{equation}
    B_{p_k}(z) =\log\left(\frac{\|p_k-z\|_2^2}{1-\|z\|_2^2}\right).
\end{equation}
The logit associated with embedding $z$ and class $k$ is then defined as 
    $\xi_k(z)=-B_{p_k}(z)+a_k,$
and class probabilities are obtained via softmax. See Supplementary Materials \ref{sec:supp_hyperbolics} for more details on hyperbolic spaces and Busemann functions. 

\subsection{Class-Incremental Semantic Segmentation}\label{sec:method_ciss}

When the label set evolves from $\mathcal{Y}_{s-1}$ to $\mathcal{Y}_s$, prototypes associated with newly introduced fine-grained categories are initialized from their parent prototypes. For a coarse class $k$ being refined into subclasses $\{k_1,...,k_m\}$, each subclass prototype $p_{k_j}$ and bias $a_{k_j}$ are initialized as \begin{equation}
\label{eq:new_prototypes}
    p_{k_j} = \frac{p_k + \epsilon_j}{\|p_k + \epsilon_j\|_2} \in \mathbb{S}^{d-1}, \qquad a_{k_j} = a_k\in\mathbb{R},
\end{equation} where $p_k\in\mathbb{S}^{d-1}$ and $a_k\in\mathbb{R}$ are the parent prototype and bias. $\epsilon_j \in\mathbb{R}^d$ is a small random perturbation drawn from a $d$-dimensional Gaussian distribution. This preserves semantic continuity: new subclasses are positioned close to their parent yet far enough to support discrimination through learning. Crucially, expansion occurs only in the prototype space, avoiding modifications of the segmentation backbone. During incremental steps, parent prototypes are frozen, and only the child prototypes and biases, as well as the backbone are learned.

\subsection{Segmentation Loss} 

To train the segmentation module, we follow previous work on hierarchical semantic segmentation \citep{li2023semantic,hindel_taxonomy-aware_2024} and leverage the class taxonomy. We frame the problem as a pixel-wise multi-label classification problem: for each pixel, the true leaf class and all its ancestors are labeled positive, while all other categories are labeled negative. We adopt a tree-min loss \citep{li2023semantic} that penalizes hierarchical inconsistencies. For each class $k\in\mathcal{Y}_s$, let $A_k$ and $D_k$ denote its ancestors and descendants in the label hierarchy (including $k)$, and let $l_k\in\{0,1\}$ indicate whether $k$ is labeled as positive. The tree-min loss is defined as: \begin{equation}
    \mathcal{L}_\mathrm{hier} = \sum_{k\in\mathcal{Y}_s} -l_k\log\left(\min_{u\in A_k}s_u\right) -(1-l_k)\log\left(1-\max_{u\in D_k}s_u\right),
\end{equation} where $s_k=\mathrm{sigmoid}(\xi_k)$. In addition to this hierarchical constraint, previous approaches \citep{hindel_taxonomy-aware_2024} use a categorical cross-entropy loss at each hierarchical level to enforce competition between sibling classes. However, since our logits are derived from Busemann distances to hyperbolic prototypes, this approach was found to lead to bias toward parent prototypes and hinder fine-grained discrimination. We instead apply cross-entropy $\mathcal{L}_\mathrm{CE}$ on active leaf classes only. The same applies during inference, where only leaf logits are considered. The final segmentation loss is a weighted combination of the tree-min and cross-entropy terms: \begin{equation}
\label{eq:loss_segm}
    \mathcal{L}_\mathrm{seg} = \beta\cdot \mathcal{L}_\mathrm{hier} + \mathcal{L}_\mathrm{CE}. 
\end{equation}

\subsection{Unsupervised Change Detection via Busemann Vector Fields}
\label{sec:method_busemann_flow}

To detect changes, we freeze the trained segmentation module and monitor differences between the hyperbolic embeddings obtained from the segmentation. Due to natural perturbations such as seasonal or lighting differences, we cannot identify changes by relying on the embeddings distance directly. Instead, we learn the natural shift within semantic classes as the hyperbolic semantic flow of a learned vector field $V$, and reconstruct the expected second embeddings ($z_2$) from the first ones ($z_1$). Remaining differences between the predicted and true embeddings will indicate change, as depicted in \cref{fig:introduction}, panel (3), framing CD as an anomaly detection problem. This approach is based on the assumption that structural changes constitute a minority that the model does not learn and fails to reconstruct.
More precisely, the predicted position of the second embeddings is obtained as $\tilde{z}_2 = \exp_{z_1}\big(V(z_1)\big)$, which corresponds to following the vector field and is the hyperbolic analog of the Euclidean update $\tilde{z}_2 = z_1 + V(z_1)$. 
To define and parameterize such a vector field $V$  in hyperbolic space,
we follow~\citet{katsmanRiemannianResidualNeural2023} and construct it based on Busemann functions and their gradients. Given any neural network $n_\omega:\mathbb R^K\to\mathbb R^K$, the vector field is defined as:
\begin{equation}
    V_{P,\omega}(z)
    =
    \sum_{k\in\mathcal{Y}_s}
    \big[n_\omega(f_P(z))\big]_k \,
    \nabla B_{p_k}(z),
\end{equation}
where $f_P(z) = \big(B_{p_1}(z), \dots, B_{p_{K_t}}(z)\big)$ and $\nabla$ is the Riemannian gradient. More details are in the Supplementary Materials \ref{sec:appendix-bvf}. While $\{p_1, \dots, p_{K_s}\}$ could in principle be any set of (learned or fixed) prototypes (See Appendix \ref{sec:appendix_other_prototypes}), we reuse the segmentation prototypes and keep them fixed.
This vector field is trained separately from the segmentation task, in an unsupervised way, with a loss corresponding to the mean reconstruction error:
\begin{equation}
    \mathcal{L}_{CD} = d\big(z_2,\ \exp_{z_1}(V_{P, \omega}(z_1))\big),
\end{equation}
where $d$ is the hyperbolic distance, see Supplementary Materials \ref{sec:supp_hyperbolic}. 
We predict change when
\begin{equation}
    {d\big(z_2,\ \exp_{z_1}(V_{P, \omega}(z_1))\big) > d_\mathrm{thresh}}, 
\end{equation} where $d_\mathrm{thresh}$ is computed using the triangle method \citep{trianglethreshold}, see Supplementary Materials \ref{sec:supp_influence_threshold} for details and values.
During incremental steps, when new prototypes are introduced, the vector field is relearned from scratch to capture transitions involving newly introduced classes.
\section{Datasets}
\label{sec:datasets}
\subsection{HiChange} 
The analysis of hierarchical and incremental change detection requires the creation of a specialized dataset that accurately represents hierarchical relations within remote sensing images, as existing SCD datasets are limited to a small number of generic and static classes. To address this gap, we introduce HiChange, a novel dataset specifically designed for hierarchical and incremental semantic change detection in aerial imagery. HiChange is constructed from \href{https://geoservices.ign.fr/bdortho}{BDOrtho} aerial imagery. 
The dataset comprises $1000$ RGB-NIR tiles, each of size $5000\times5000$ pixels at $0.2$m spatial resolution, sampled over France, and partitioned geographically into training, validation, and test sets ($70\%-10\%-20\%$), ensuring no spatial overlap across splits.  
HiChange is fully annotated following a hierarchical taxonomy. Coarse semantic information derived from \href{https://geoservices.ign.fr/ocsge}{OCSGE} 
is combined with detailed labels from FLAIR \citep{garioud2025flair} and \href{https://geoservices.ign.fr/bdtopo}{BDTopo} 
to label each image at different semantic levels, for which the user can specify between $4$ and $32$ classes of interest.
Additional information on the dataset, including the complete hierarchical taxonomy, details on the label generation pipeline and the sampling strategy as well as statistics and comparison with existing benchmarks, is provided in the Supplementary Material \ref{sec:supp_hichange}. 

\subsection{Additional Datasets}

While HiChange is the only available dataset with sufficient semantic richness to support hierarchical CI-SCD, we complement our evaluation with experiments on two widely used SCD benchmarks: SECOND \citep{yang2020semantic} and Hi-UCD-mini \citep{tian2020hi}.
Although these datasets do not support incremental evaluation, they allow us to validate the segmentation backbone and Busemann vector field–based change detection mechanism, demonstrating that our approach generalizes beyond HiChange.
More details on these datasets can be found in Supplementary Materials \ref{sec:supp_other_ds}. 
\section{Experiments}
\label{sec:results}

\begin{figure}[t]
    \centering
    \includegraphics[width=\linewidth]{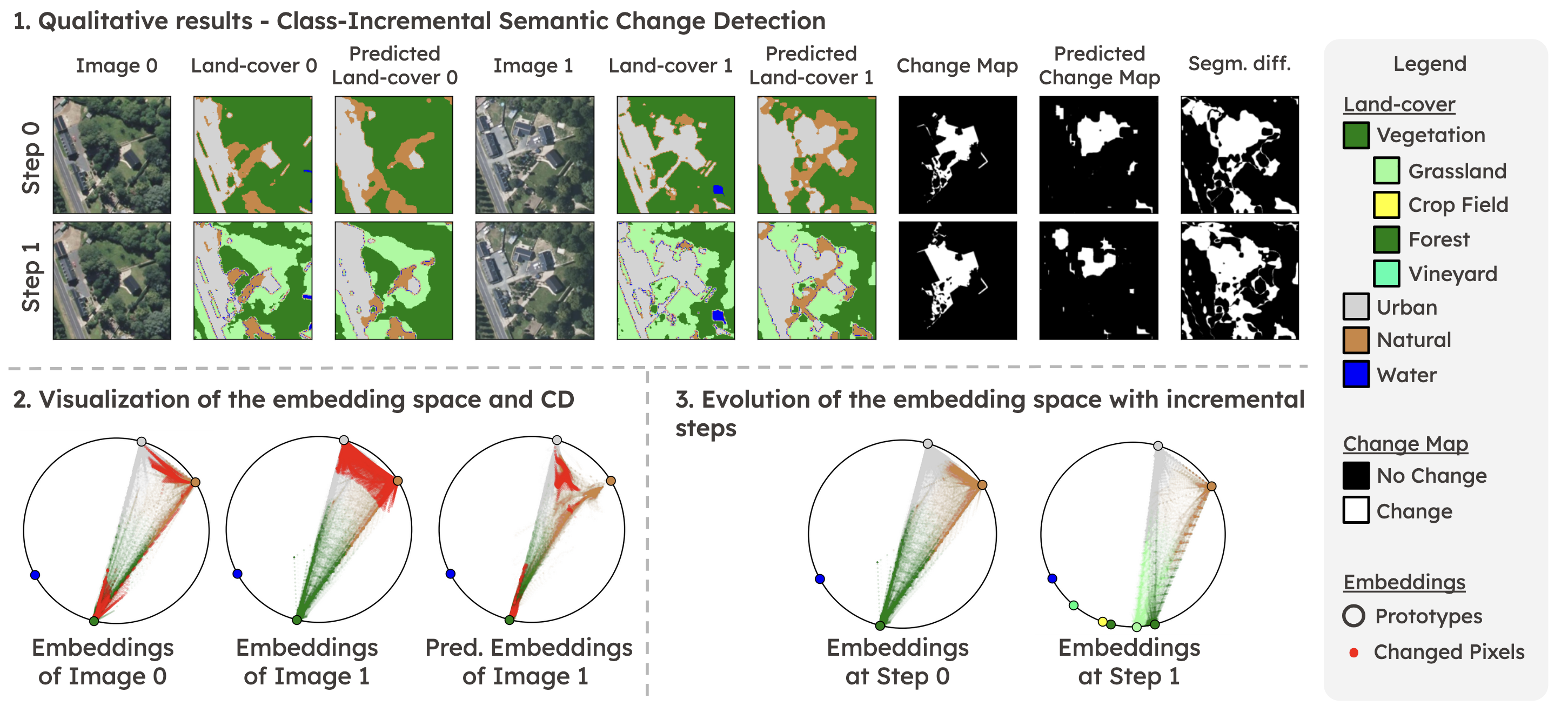}
    \caption{Qualitative illustration of HyperFlow. (1) SCD at two incremental steps. For each step, we show segmentation results at both timestamps, predicted change map, and a naive baseline obtained by differencing the two segmentation maps (Segm. diff.). (2) Hyperbolic 2-$d$ embeddings of the first and second image, and predicted second embeddings obtained via the learned vector field. Pixels corresponding to true changes (in red) exhibit significant displacement between the timestamps, while unchanged regions show minor displacements. (3) Evolution of the embedding space across incremental steps. We visualize embeddings for a single image before and after the introduction of new classes. More qualitative examples are provided in Supplementary Materials \textit{S5}.}
    \label{fig:qualitative_results}
\end{figure}

\subsection{Does Hyperbolic Semantic Flow Enable Effective CD?}
\label{sec:results_part1}
We first 
compare HyperFlow against unsupervised CD baselines across the three datasets. We also report the performance of the fully supervised SCD method HRSCD4 \citep{daudt2019multitask} as an upper-bound reference. Implementation details are given in the Supplementary Materials \ref{sec:supp_implementation}, and qualitative examples are provided in \cref{fig:qualitative_results}, panel (1). Numerically, \cref{tab:CD_perfs} shows that HyperFlow consistently outperforms competing approaches across all datasets.
To isolate the contribution of the learned vector field, we additionally compare against a naive baseline that defines change as the difference between the segmentation of the two input images (Segm. diff.). Although this baseline already surpasses several unsupervised methods, and thus suggests that strong semantic segmentation alone can outperform many unsupervised CD approaches, an important performance gap remains with  HyperFlow. This difference highlights that in the naive baseline small variations in appearance between the images can slightly modify the semantic embeddings, potentially leading to label changes in the output even when no real semantic change occurred. In contrast, HyperFlow operates directly in the embedding space and measures deviations of a semantic flow accounting for these variations. These irrelevant changes are therefore not accounted for, and only deviations that cannot be explained by the learned semantic flow, \ie, semantically significant movement in the embedding space, are identified as changes.
This mechanism is illustrated in \cref{fig:qualitative_results}, panel (2): only pixels corresponding to true changes (shown in red) undergo significant displacement between timestamps. The hyperbolic flow therefore controls segmentation-induced errors, explaining the improvement over simple segmentation differencing, also visible in \cref{fig:qualitative_results}, panel (1). 
\begin{table}[t!]
\centering
\caption{Comparison of HyperFlow with state-of-the-art unsupervised CD methods across the three datasets, for the task $T_{K_0}^0$, \ie before any incremental step is performed. For HyperFlow, we additionally report CD performance using different embedding sources: HyperFlow uses the in-domain embeddings produced by the segmentation pipeline, while HyperFlow (SAM3, AnySat) use embeddings from the corresponding foundation models. Bold and underline indicate the first and second-best unsupervised scores, respectively. $\dagger$ The HRSCD4 model is supervised.}
\label{tab:CD_perfs}
\renewcommand{\tabcolsep}{4pt}
\begin{threeparttable}
\resizebox{\textwidth}{!}{\begin{tabular}{@{}ccccc|cccc|cccc@{}}
\toprule
\multicolumn{1}{c}{}                      &\multicolumn{4}{c}{\textbf{HiChange}}                                                                                                         & \multicolumn{4}{c}{\textbf{SECOND}}                                                                                                         & \multicolumn{4}{c}{\textbf{Hi-UCD-mini}}                                                                                                     \\            & \textbf{Acc.}                     & \textbf{F1}                       & \textbf{Pre.}                     & \textbf{Rec.}                     & \multicolumn{1}{c}{\textbf{Acc.}} & \multicolumn{1}{c}{\textbf{F1}} & \multicolumn{1}{c}{\textbf{Pre.}} & \multicolumn{1}{c|}{\textbf{Rec.}} & \multicolumn{1}{c}{\textbf{Acc.}} & \multicolumn{1}{c}{\textbf{F1}} & \multicolumn{1}{c}{\textbf{Pre.}} & \multicolumn{1}{c}{\textbf{Rec.}} \\ \midrule
\textbf{CVA~\citep{malila1980change}}                      & 74.2                              & 8.0                              & 4.5                              & 39.3                              &        67.6                           &                    30.5             &    25.9                              &   37.1                                &       75.5                            &   14.4                              &       9.3                           &        32.3                           \\
\textbf{SiROC~\citep{kondmann2021spatial}}                         & 78.0                              & 9.3                              & 5.3                            & 39.1                             &          69.4                         &                  29.6               &     26.5                              &    33.4                               &         74.4                         &      15.6                           &        9.9                           &                 36.9                  \\

\textbf{CDRL~\citep{noh2022unsupervised}}                            & 73.0                              & 11.6                              & 6.4 & 61.9                              &     62.3                              &    33.2                            &      25.2                            &         48.8                          &                  81.6                 &         17.9                        &    12.5                               &      31.3                             \\
\textbf{X-Net~\citep{luppino2021deep}}                                 & 69.0                            & 11.5                              & 6.3                              & \underline{70.1}                              &           60.8                        &                      34.7           &       25.5                            &    54.2                     &               84.9                   &                 20.4               &  15.4                                 &   30.1                              \\
\textbf{ACE-Net~\citep{luppino2021deep}}                        & 69.9                              & 11.0                              & 6.0                              & 65.0                              &                65.7                   &                33.2                 &    26.5                              &       44.4                            &     78.1                              &            20.0                     &    13.1                              &     42.6                              \\
\textbf{KPCAMNet~\citep{wu2021unsupervised}}                         & 92.5                              & 20.2                              & 14.6                              & 33.2                              &  \underline{77.9}                      &                 24.4                &        35.6                           &       18.6                            &    \underline{89.2}                               &       16.8                         &       16.7                           &                    17.0               \\
\textbf{AnyChange~\citep{zheng2024segment}}               & \textbf{94.0}                     & 19.4 & \underline{15.8}                    & 25.1                              &     \textbf{81.2}                              &   33.0                              &                        \textbf{52.3}           &      24.1                            &                      88.2             &     26.5                            &    \underline{22.1}            &            33.1             \\
\textbf{2Player~\citep{bechaz20262player}}                          & 92.0                             & 17.1                              & 12.2                              & 28.7                             &  69.0                                 &                 34.1               &                         28.6        &                       42.1            &        76.7                           &    21.4                             &     13.7                              &     48.6                              \\ \midrule
\textbf{Segm. diff.} &  86.7                     &              \underline{25.1}          & {14.9} & \textbf{78.8} &          60.9                         &             \underline{41.9}                 &         29.3                       &                 73.3              &                73.3             &                \underline{28.1}               &       17.0                    &             \underline{80.8}                     \\
\textbf{HyperFlow}&    92.9 &   \textbf{29.6}              &   \textbf{20.5}                   &   52.8                         &      73.7                       &            \textbf{43.0}                &        \underline{36.8}                   &        51.8                   &   \textbf{91.7}                    &   \textbf{44.3}                    &                \textbf{39.0}        &               51.4                    \\

\textbf{HyperFlow (SAM3)}&    \underline{93.9} &   15.1            &   12.3                  &   19.6                      &    48.5        &      40.3        &            25.9            &           \textbf{90.8}                       &            49.7           &         18.3             &         10.2             &   \textbf{88.0}  \\
\textbf{HyperFlow (AnySat)}&    93.5 &   20.1              &   14.2                  &   34.5                         &      45.0                    &       36.5                   &        23.4                 &             \underline{82.2}         &           84.3           &     23.7                &             17.4          &         37.1                       \\

\midrule
\textbf{HRSCD4$\dagger$~\citep{daudt2019multitask}}  &  94.2   &        42.9        &    29.8                &            76.4              &          66.5                    &     52.0                       &     35.9                     &                       94.4        &                    94.0          &                   60.9           &         52.3                         &           72.9                      \\

\bottomrule
\end{tabular}}

\end{threeparttable}
\end{table} 


When compared with existing unsupervised methods, the strong CD results in \cref{tab:CD_perfs} may raise concerns about the unsupervised nature of HyperFlow, given that not only the prototypes but also the embeddings are obtained from supervised segmentation, therefore influencing the performance of Hyperflow. We emphasize that our unsupervised claim concerns the CD stage only. For CD, the supervised segmentation backbone acts purely as a feature extractor, and in this sense, Hyperflow follows common practices in unsupervised CD of using supervised pretrained feature extractors \citep{saha2019unsupervised,lei2026remote,bergamasco2022unsupervised,du2018unsupervised,zheng2024segment}. HyperFlow differs in that the features are specifically adapted to the target domain and semantic task, thus yielding stronger features for CD, but HyperFlow could also operate on embeddings obtained through either supervised or unsupervised pretraining, with their quality and task/domain specificity naturally affecting downstream CD performance. 
To support this statement, \cref{tab:CD_perfs} also reports results using embeddings from two foundation models: SAM3 \citep{carion2026sam3segmentconcepts} and AnySat \citep{astruc2025anysat}, the latter being RS-specific. Even with these fully unsupervised embeddings, HyperFlow achieves performance comparable to SOTA unsupervised CD methods. Moreover, for HiChange and Hi-UCD-mini, performance improves from SAM3 to AnySat, consistent with increasing specialization to the RS domain. Thus, while supervised segmentation embeddings contribute to the strong results observed with HyperFlow, they are not required for competitive unsupervised CD performance. More generally, HyperFlow benefits from strong representations, regardless of how they are obtained. 

\subsection{Class-incremental settings}

When considering the incremental setting (\cref{tab:change_as_classes_added}), HyperFlow maintains strong SCD performance even as the number of semantic classes grows over incremental steps.
While these scenarios are inherently more challenging, not only does HyperFlow still outperform the strongest unsupervised CD baselines presented in \cref{tab:CD_perfs}, but it is also capable of characterizing the nature of a change at a much finer lever of the semantic taxonomy.
\begin{table}[h!]
\centering
\caption{SCD performance of HyperFlow across incremental steps for the HiChange dataset. We report segmentation performance with mIoU, CD with F1, and combined CD and segmentation using the Semantic Change Segmentation (SCS) score from \cite{toker2022dynamicearthnet}.}
\label{tab:change_as_classes_added}
\begin{tabular}{@{}cccccccc@{}}
\toprule
                      & $T^0_{4}$ & $T^1_{7}$ & $T^2_{13}$ & $T^3_{16}$ & $T^4_{20}$ & $T^5_{22}$ & $T^6_{32}$ \\ \midrule
\textbf{Segm. [mIoU]}   & 62.0 & 56.2 & 33.7 & 28.4 & 23.5 & 20.7 &  16.1 \\
\textbf{CD [F1]} & 29.6 & 29.6 & 29.8 & 29.2& 27.7 & 31.6 &  26.0\\
\textbf{SCD [SCS]} & 39.0 & 34.4 & 22.4 & 19.9 & 17.3 & 17.6 & 13.8 \\  \bottomrule
\end{tabular}

\end{table}

\Cref{tab:segm_perfs} compares HyperFlow segmentation results against state-of-the-art CISS methods across incremental steps for the three datasets considered. HyperFlow achieves competitive performance and, while improvements over prior approaches remain moderate, the results demonstrate that considering Busemann prototypes and incorporating CD do not degrade CISS performance. 

\begin{table}[]
\centering
\caption{Comparison of Hyperflow with state-of-the-art CISS methods for the three datasets. The mIoU is reported for each method and all incremental steps. Note that since SECOND and Hi-UCD do not enable incremental scenarios, only the base task is considered (Hyp: hyperbolic).}
\label{tab:segm_perfs}
\renewcommand{\tabcolsep}{3pt} 
\resizebox{\textwidth}{!}{\begin{tabular}{@{}cc|ccccccc|c|c@{}}
\toprule
                   &                     & \multicolumn{7}{c|}{\textbf{HiChange}}                                                                & \textbf{SECOND} & \textbf{Hi-UCD} \\
\textbf{Method}    & \textbf{Hyp.} & $T^0_{4}$ & $T^1_{7}$ & $T^2_{13}$ & $T^3_{16}$ & $T^4_{20}$ & $T^5_{22}$ & $T^6_{32}$ & $T^0_{6}$      & $T^0_{9}$          \\ \midrule
\textbf{PLOP~\citep{douillard2021plop}}      &      \ding{55}               &     \underline{66.3}       &  11.4   &    2.0      &    0.4    &     0.9       &    0.8       &   0.6      &       \textbf{53.7}         &        41.8             \\
\textbf{MiB~\citep{cermelli2020modeling}}       &    \ding{55}                 &    \underline{66.3}       &   52.9     &     \underline{36.0}      &    27.8       &     22.2        &     \underline{20.7}       &     \underline{16.1}        &        \textbf{53.7}        &       41.8               \\
\textbf{MiB+AWT~\citep{goswami2023attribution}} &   \ding{55}                  &       \underline{66.3}     &      54.7     &     \textbf{37.6}     &     \underline{28.2}       &       \underline{22.9}    &     20.2  
&    15.9        &        \textbf{53.7}        &            41.8          \\
\textbf{DKD~\citep{baek2022decomposed}}       &   \ding{55}                  &      64.7      &    \textbf{57.4}      &    35.0      &    27.8     &   22.1         &    \textbf{21.5}        &    \textbf{17.2}         &       38.8          &         \textbf{45.8}            \\
\textbf{TOPICS~\citep{hindel_taxonomy-aware_2024}}    &    \ding{51}                 &     \textbf{67.6}       &            54.5      &    24.6        &      15.9      &    8.5     &  6.2   &  3.8  &    46.8             &     37.5                \\
\textbf{SAM3~\citep{carion2026sam3segmentconcepts}}    &    \ding{55}                 &      29.8      &       24.3       &    20.5     &   13.8         &       11.2     &     9.3    &   6.4     &      28.0          &        20.9            \\
\midrule
\textbf{HyperFlow}  &  \ding{51}                   &    62.0 & \underline{56.2} & 33.7 & \textbf{28.4} & \textbf{23.5} & \underline{20.7} &  \underline{16.1}       &           \underline{47.3}      &           \underline{42.4}          \\ \bottomrule
\end{tabular}}

\end{table}


We also investigate whether progressively refining the taxonomy provides advantages over directly training on the full set of fine-grained categories. 
Results in \cref{tab:incr_vs_non_incr2} show that structuring semantic representations progressively, rather than directly considering fine-grained distinctions, leads to improved SCD. 
Figure \ref{fig:qualitative_results}, panel (3) visualizes this effect, showing how incremental refinement preserves global organization while progressively increasing semantic resolution. More embeddings comparisons across incremental and non-incremental scenarios are presented in the Supplementary Materials \ref{sec:supp_qualitative}. 

To isolate the effect of hierarchical initialization, we also investigate in \cref{tab:incr_vs_non_incr2} a non-incremental scenario where prototypes are initialized according to the class hierarchy. Results improve compared to the non-incremental baseline, indicating that respecting class hierarchies at initialization helps segmentation. Performance remains however lower than in the full incremental setting with progressive taxonomy refinement, showing that benefits of incremental learning go beyond hierarchical initialization alone.

\begin{table}[h]
\centering
\caption{Comparison between (i) class-incremental training with progressive taxonomy refinement, (ii) non-incremental training on the full fine-grained taxonomy, and (iii) non-incremental training with hierarchical prototype initialization for the HiChange dataset. The SCS score is reported for all incremental steps. At each step, the incremental and non incremental models are trained for the same total number of epochs.}
\label{tab:incr_vs_non_incr2}
\begin{tabular}{ccccccccc}
\toprule
\textbf{Incr.}             & \textbf{Hier.}             & $T^0_{4}$ & $T^1_{7}$ & $T^2_{13}$ & $T^3_{16}$ & $T^4_{20}$ & $T^5_{22}$ & $T^6_{32}$ \\ \midrule
\ding{51} & \ding{51} &      \textbf{39.0} & \textbf{34.4} & \textbf{22.4}& \textbf{19.9} & \textbf{17.3} & \textbf{17.6} & \textbf{13.8}       \\
\ding{55} & \ding{55} &    \textbf{39.0}        &    30.9      &      16.6       &    13.7         &     11.7        &      11.7      &      9.1      \\
\ding{55} & \ding{51} &    \textbf{39.0}        &     \underline{31.7}       &       \underline{17.8}      &     \underline{16.2}         &      \underline{14.4}       &     \underline{14.1}        &     \underline{10.0}        \\ \bottomrule
\end{tabular}
\end{table}



\section{Conclusion}
\label{sec:discussion}
In this work, we introduce HyperFlow, a method that bridges CISS and SCD via hyperbolic semantic flow. By representing semantic categories with Busemann prototypes and modeling temporal evolution of semantic classes as a learned vector field in hyperbolic space, we formulate CD as the identification of deviations from expected temporal dynamics. We train and evaluate HyperFlow on HiChange, a new large-scale, very high resolution and hierarchical SCD dataset that we propose. Our results demonstrate that formulating semantic change as a hyperbolic flow enables effective unsupervised CD while preserving competitive segmentation performance, both in standard and in incremental settings. Furthermore, we show that progressively refining the taxonomy, \ie the semantic changes considered, rather than considering all changes from the beggining, helps considerably the task of fine-grained semantic change detection. Although the CD mechanism does not require supervision, HyperFlow still depends on the quality of the semantic segmentation model. Future works will focus on learning better representations for the semantic segmentation backbone, for instance leveraging foundation models based on large scale self-supervised pretraining \citep{sumbul2025smarties,cong2022satmae}. Moreover, we will explore leveraging longer time series, which could enable the modeling of more complex or gradual temporal transitions, such as seasonal vegetation cycles or multi-stage urban development, thereby providing a richer representation of temporal dynamics.

\subsection*{AI use statement}

In this work, we used generative AI tools to assist with debugging and improving existing code, and to reformulate some paragraphs of the manuscript to improve readability. The original code and scientific text were however written by the authors and were not generated by generative AI. Generative AI tools were not used to generate research ideas or design the methodology and experiments, nor to interpret the results. All AI-generated modifications were carefully reviewed by the authors. Code modifications were verified for correctness, and changes to the manuscript carefully reviewed by multiple authors to ensure that they accurately reflected the intended meaning. We take full responsibility for the final content of this work. 


\subsection*{Reproducibility statement}

To facilitate reproducibility, all code used in our experiments will be made publicly available upon acceptance of the paper. Detailed information on the generation of the HiChange dataset is provided in the supplementary material \ref{sec:supp_hichange}. The images and label sources used to construct the HiChange dataset are publicly available, with their respective sources and links provided in Appendix \ref{sec:supp_hichange}. The code used to fuse the different label sources will also be published. The resulting HiChange dataset will be publicly released upon acceptance of the paper. All other datasets used in our experiments are publicly available, and their sources are provided in Appendix \ref{sec:supp_other_ds}. Implementation details are provided in Appendix \ref{sec:supp_implementation}. Pretrained model checkpoints used to obtain the reported results will also be released alongside the source code.



\bibliography{main}

@String(CVPR  = {IEEE Conf. Comput. Vis. Pattern Recog.})

@String(BMVC  = {Brit. Mach. Vis. Conf.})

@String(AAAI  = {AAAI})

@String(ICIP  = {IEEE Int. Conf. Image Process.})

@String(ICASSP=	{ICASSP})

@String(CVPR  = {CVPR})

@String(BMVC  =	{BMVC})

@String(ICIP  = {ICIP})

@article{wang2024rsbuilding,
  title={Rsbuilding: Towards general remote sensing image building extraction and change detection with foundation model},
  author={Wang, Mingze and Su, Lili and Yan, Cilin and Xu, Sheng and Yuan, Pengcheng and Jiang, Xiaolong and Zhang, Baochang},
  journal={IEEE Transactions on Geoscience and Remote Sensing},
  year={2024},
  publisher={IEEE}
}

@article{zhao2025df2net,
  title={DF2Net: A Differential-Feature Dynamic-Fusion Network for Building Change Detection in Remote Sensing Images},
  author={Zhao, Like and Yang, Mingwang and Zhao, Yiran and Liu, Yuqing and Jiang, Huawei and Yang, Zhen},
  journal={IEEE Transactions on Geoscience and Remote Sensing},
  volume={64},
  pages={1--15},
  year={2025},
  publisher={IEEE}
}

@inproceedings{daudt2018fully,
  title={Fully convolutional siamese networks for change detection},
  author={Daudt, Rodrigo Caye and Le Saux, Bertr and Boulch, Alexandre},
  booktitle={2018 25th IEEE international conference on image processing (ICIP)},
  pages={4063--4067},
  year={2018},
  organization={IEEE}
}

@article{zheng2021building,
  title={Building damage assessment for rapid disaster response with a deep object-based semantic change detection framework: From natural disasters to man-made disasters},
  author={Zheng, Zhuo and Zhong, Yanfei and Wang, Junjue and Ma, Ailong and Zhang, Liangpei},
  journal={Remote Sensing of Environment},
  volume={265},
  pages={112636},
  year={2021},
  publisher={Elsevier}
}

@inproceedings{ahn2025generalizable,
  title={Generalizable disaster damage assessment via change detection with vision foundation model},
  author={Ahn, Kyeongjin and Han, Sungwon and Park, Sungwon and Kim, Jihee and Park, Sangyoon and Cha, Meeyoung},
  booktitle={Proceedings of the AAAI Conference on Artificial Intelligence},
  volume={39},
  number={27},
  pages={27784--27792},
  year={2025}
}

@article{li2016superresolution,
  title={A superresolution land-cover change detection method using remotely sensed images with different spatial resolutions},
  author={Li, Xiaodong and Ling, Feng and Foody, Giles M and Du, Yun},
  journal={IEEE Transactions on Geoscience and Remote Sensing},
  volume={54},
  number={7},
  pages={3822--3841},
  year={2016},
  publisher={IEEE}
}

@article{jiang2022survey,
  title={A survey on deep learning-based change detection from high-resolution remote sensing images},
  author={Jiang, Huiwei and Peng, Min and Zhong, Yuanjun and Xie, Haofeng and Hao, Zemin and Lin, Jingming and Ma, Xiaoli and Hu, Xiangyun},
  journal={Remote Sensing},
  volume={14},
  number={7},
  pages={1552},
  year={2022},
  publisher={MDPI}
}

@article{lv2023spatial,
  title={Spatial-contextual information utilization framework for land cover change detection with hyperspectral remote sensed images},
  author={Lv, Zhiyong and Zhang, Ming and Sun, Weiwei and Benediktsson, J{\'o}n Atli and Lei, Tao and Falco, Nicola},
  journal={IEEE Transactions on Geoscience and Remote Sensing},
  volume={61},
  pages={1--11},
  year={2023},
  publisher={IEEE}
}

@article{daudt2019multitask,
  title={Multitask learning for large-scale semantic change detection},
  author={Daudt, Rodrigo Caye and Le Saux, Bertrand and Boulch, Alexandre and Gousseau, Yann},
  journal={Computer Vision and Image Understanding},
  volume={187},
  pages={102783},
  year={2019},
  publisher={Elsevier}
}

@article{touati2020anomaly,
  title={Anomaly feature learning for unsupervised change detection in heterogeneous images: A deep sparse residual model},
  author={Touati, Redha and Mignotte, Max and Dahmane, Mohamed},
  journal={IEEE Journal of Selected Topics in Applied Earth Observations and Remote Sensing},
  volume={13},
  pages={588--600},
  year={2020},
  publisher={IEEE}
}

@article{andresini2022leveraging,
  title={Leveraging autoencoders in change vector analysis of optical satellite images},
  author={Andresini, Giuseppina and Appice, Annalisa and Iaia, Daniele and Malerba, Donato and Taggio, Nicolo and Aiello, Antonello},
  journal={Journal of Intelligent Information Systems},
  volume={58},
  number={3},
  pages={433--452},
  year={2022},
  publisher={Springer}
}

@article{bechaz20262player,
  title={2Player: A general framework for self-supervised change detection via cooperative learning},
  author={B{\'e}chaz, Manon and Dalsasso, Emanuele and Tomoiag{\u{a}}, Ciprian and Detyniecki, Marcin and Tuia, Devis},
  journal={ISPRS Journal of Photogrammetry and Remote Sensing},
  volume={232},
  pages={34--47},
  year={2026},
  publisher={Elsevier}
}

@article{ruuvzivcka2022ravaen,
  title={RaV{\AE}n: unsupervised change detection of extreme events using ML on-board satellites},
  author={Růžička, V{\'\i}t and Vaughan, Anna and De Martini, Daniele and Fulton, James and Salvatelli, Valentina and Bridges, Chris and Mateo-Garcia, Gonzalo and Zantedeschi, Valentina},
  journal={Scientific reports},
  volume={12},
  number={1},
  pages={16939},
  year={2022},
  publisher={Nature Publishing Group UK London}
}

@inproceedings{noh2022unsupervised,
  title={Unsupervised change detection based on image reconstruction loss},
  author={Noh, Hyeoncheol and Ju, Jingi and Seo, Minseok and Park, Jongchan and Choi, Dong-Geol},
  booktitle={proceedings of the IEEE/CVF conference on computer vision and pattern recognition},
  pages={1352--1361},
  year={2022}
}

@article{liu2021unsupervised,
  title={Unsupervised change detection from heterogeneous data based on image translation},
  author={Liu, Zhun-Ga and Zhang, Zuo-Wei and Pan, Quan and Ning, Liang-Bo},
  journal={IEEE Transactions on Geoscience and Remote Sensing},
  volume={60},
  pages={1--13},
  year={2021},
  publisher={IEEE}
}

@article{luppino2021deep,
  title={Deep image translation with an affinity-based change prior for unsupervised multimodal change detection},
  author={Luppino, Luigi Tommaso and Kampffmeyer, Michael and Bianchi, Filippo Maria and Moser, Gabriele and Serpico, Sebastiano Bruno and Jenssen, Robert and Anfinsen, Stian Normann},
  journal={IEEE Transactions on Geoscience and Remote Sensing},
  volume={60},
  pages={1--22},
  year={2021},
  publisher={IEEE}
}

@article{huang2010sampling,
  title={Sampling approaches for one-pass land-use/land-cover change mapping},
  author={Huang, Zhi and Jia, Xiuping and Ge, Linlin},
  journal={International Journal of Remote Sensing},
  volume={31},
  number={6},
  pages={1543--1554},
  year={2010},
  publisher={Taylor \& Francis}
}

@article{hu2013seasonal,
  title={Seasonal change of land-use/land-cover (LULC) detection using MODIS data in rapid urbanization regions: A case study of the pearl river delta region (China)},
  author={Hu, Jinrong and Zhang, Yuanzhi},
  journal={IEEE Journal of Selected Topics in Applied Earth Observations and Remote Sensing},
  volume={6},
  number={4},
  pages={1913--1920},
  year={2013},
  publisher={IEEE}
}

@article{wu2017post,
  title={A post-classification change detection method based on iterative slow feature analysis and Bayesian soft fusion},
  author={Wu, Chen and Du, Bo and Cui, Xiaohui and Zhang, Liangpei},
  journal={Remote Sensing of Environment},
  volume={199},
  pages={241--255},
  year={2017},
  publisher={Elsevier}
}

@article{zhu2022land,
  title={Land-use/land-cover change detection based on a Siamese global learning framework for high spatial resolution remote sensing imagery},
  author={Zhu, Qiqi and Guo, Xi and Deng, Weihuan and Shi, Sunan and Guan, Qingfeng and Zhong, Yanfei and Zhang, Liangpei and Li, Deren},
  journal={ISPRS Journal of Photogrammetry and Remote Sensing},
  volume={184},
  pages={63--78},
  year={2022},
  publisher={Elsevier}
}

@article{ding2024joint,
  title={Joint spatio-temporal modeling for semantic change detection in remote sensing images},
  author={Ding, Lei and Zhang, Jing and Guo, Haitao and Zhang, Kai and Liu, Bing and Bruzzone, Lorenzo},
  journal={IEEE Transactions on Geoscience and Remote Sensing},
  volume={62},
  pages={1--14},
  year={2024},
  publisher={IEEE}
}

@article{cha2021ssul,
  title={Ssul: Semantic segmentation with unknown label for exemplar-based class-incremental learning},
  author={Cha, Sungmin and Yoo, YoungJoon and Moon, Taesup and others},
  journal={Advances in neural information processing systems},
  volume={34},
  pages={10919--10930},
  year={2021}
}

@article{chen2023saving,
  title={Saving 100x storage: Prototype replay for reconstructing training sample distribution in class-incremental semantic segmentation},
  author={Chen, Jinpeng and Cong, Runmin and Luo, Yuxuan and Ip, Horace and Kwong, Sam},
  journal={Advances in Neural Information Processing Systems},
  volume={36},
  pages={35988--35999},
  year={2023}
}

@inproceedings{maracani2021recall,
  title={Recall: Replay-based continual learning in semantic segmentation},
  author={Maracani, Andrea and Michieli, Umberto and Toldo, Marco and Zanuttigh, Pietro},
  booktitle={Proceedings of the IEEE/CVF international conference on computer vision},
  pages={7026--7035},
  year={2021}
}

@inproceedings{michieli2019incremental,
  title={Incremental learning techniques for semantic segmentation},
  author={Michieli, Umberto and Zanuttigh, Pietro},
  booktitle={Proceedings of the IEEE/CVF international conference on computer vision workshops},
  pages={0--0},
  year={2019}
}

@article{baek2022decomposed,
  title={Decomposed knowledge distillation for class-incremental semantic segmentation},
  author={Baek, Donghyeon and Oh, Youngmin and Lee, Sanghoon and Lee, Junghyup and Ham, Bumsub},
  journal={Advances in neural information processing systems},
  volume={35},
  pages={10380--10392},
  year={2022}
}

@article{srivastava2019adaptive,
  title={Adaptive compression-based lifelong learning},
  author={Srivastava, Shivangi and Berman, Maxim and Blaschko, Matthew B and Tuia, Devis},
  journal={arXiv preprint arXiv:1907.09695},
  year={2019}
}

@inproceedings{zhang2022representation,
  title={Representation compensation networks for continual semantic segmentation},
  author={Zhang, Chang-Bin and Xiao, Jia-Wen and Liu, Xialei and Chen, Ying-Cong and Cheng, Ming-Ming},
  booktitle={Proceedings of the IEEE/CVF Conference on Computer Vision and Pattern Recognition},
  pages={7053--7064},
  year={2022}
}

@inproceedings{xiao2023endpoints,
  title={Endpoints weight fusion for class incremental semantic segmentation},
  author={Xiao, Jia-Wen and Zhang, Chang-Bin and Feng, Jiekang and Liu, Xialei and van de Weijer, Joost and Cheng, Ming-Ming},
  booktitle={Proceedings of the IEEE/CVF conference on computer vision and pattern recognition},
  pages={7204--7213},
  year={2023}
}

@inproceedings{cermelli2020modeling,
  title={Modeling the background for incremental learning in semantic segmentation},
  author={Cermelli, Fabio and Mancini, Massimiliano and Bulo, Samuel Rota and Ricci, Elisa and Caputo, Barbara},
  booktitle={Proceedings of the IEEE/CVF conference on computer vision and pattern recognition},
  pages={9233--9242},
  year={2020}
}

@inproceedings{zhu2025adaptive,
  title={Adaptive prototype replay for class incremental semantic segmentation},
  author={Zhu, Guilin and Wu, Dongyue and Gao, Changxin and Wang, Runmin and Yang, Weidong and Sang, Nong},
  booktitle={Proceedings of the AAAI Conference on Artificial Intelligence},
  volume={39},
  number={10},
  pages={10932--10940},
  year={2025}
}

@inproceedings{khrulkov_hyperbolic_2020,
  title={Hyperbolic image embeddings},
  author={Khrulkov, Valentin and Mirvakhabova, Leyla and Ustinova, Evgeniya and Oseledets, Ivan and Lempitsky, Victor},
  booktitle={Proceedings of the IEEE/CVF conference on computer vision and pattern recognition},
  pages={6418--6428},
  year={2020}
}

@inproceedings{liu_hyperbolic_2020,
  title={Hyperbolic visual embedding learning for zero-shot recognition},
  author={Liu, Shaoteng and Chen, Jingjing and Pan, Liangming and Ngo, Chong-Wah and Chua, Tat-Seng and Jiang, Yu-Gang},
  booktitle={Proceedings of the IEEE/CVF conference on computer vision and pattern recognition},
  pages={9273--9281},
  year={2020}
}

@inproceedings{lang_hyperbolic_2022,
  title={On hyperbolic embeddings in object detection},
  author={Lang, Christopher and Braun, Alexander and Schillingmann, Lars and Valada, Abhinav},
  booktitle={DAGM German Conference on Pattern Recognition},
  pages={462--476},
  year={2022},
  organization={Springer}
}

@inproceedings{ghadimiatigh_hyperbolic_2022,
  title={Hyperbolic image segmentation},
  author={Ghadimi Atigh, Mina and Schoep, Julian and Acar, Erman and Van Noord, Nanne and Mettes, Pascal},
  booktitle={Proceedings of the IEEE/CVF conference on computer vision and pattern recognition},
  pages={4453--4462},
  year={2022}
}

@article{hindel_taxonomy-aware_2024,
  title={Taxonomy-aware continual semantic segmentation in hyperbolic spaces for open-world perception},
  author={Hindel, Julia and Cattaneo, Daniele and Valada, Abhinav},
  journal={IEEE Robotics and Automation Letters},
  volume={10},
  number={2},
  pages={1904--1911},
  year={2024},
  publisher={IEEE}
}

@inproceedings{berg2024horospherical,
  title={Horospherical Learning with Smart Prototypes},
  author={Berg, Paul and Michele, Bjoern and Pham, Minh-Tan and Chapel, Laetitia and Courty, Nicolas},
  booktitle={British Machine Vision Conference (BMVC)},
  year={2024}
}

@article{cui_rethinking_2024,
  title={Rethinking few-shot class-incremental learning with open-set hypothesis in hyperbolic geometry},
  author={Cui, Yawen and Yu, Zitong and Peng, Wei and Tian, Qi and Liu, Li},
  journal={IEEE Transactions on Multimedia},
  volume={26},
  pages={5897--5910},
  year={2023},
  publisher={IEEE}
}

@ARTICLE{ostu,
  author={Otsu, Nobuyuki},
  journal={IEEE Transactions on Systems, Man, and Cybernetics}, 
  title={A Threshold Selection Method from Gray-Level Histograms}, 
  year={1979},
  volume={9},
  number={1},
  pages={62-66},
  doi={10.1109/TSMC.1979.4310076}}

@article{trianglethreshold,
  title={Automatic measurement of sister chromatid exchange frequency.},
  author={Zack, Gregory W and Rogers, William E and Latt, Samuel A},
  journal={Journal of Histochemistry \& Cytochemistry},
  volume={25},
  number={7},
  pages={741--753},
  year={1977},
  publisher={SAGE Publications Sage CA: Los Angeles, CA}
}

@misc{carion2026sam3segmentconcepts,
      title={SAM 3: Segment Anything with Concepts}, 
      author={Nicolas Carion and Laura Gustafson and Yuan-Ting Hu and Shoubhik Debnath and Ronghang Hu and Didac Suris and Chaitanya Ryali and Kalyan Vasudev Alwala and Haitham Khedr and Andrew Huang and Jie Lei and Tengyu Ma and Baishan Guo and Arpit Kalla and Markus Marks and Joseph Greer and Meng Wang and Peize Sun and Roman Rädle and Triantafyllos Afouras and Effrosyni Mavroudi and Katherine Xu and Tsung-Han Wu and Yu Zhou and Liliane Momeni and Rishi Hazra and Shuangrui Ding and Sagar Vaze and Francois Porcher and Feng Li and Siyuan Li and Aishwarya Kamath and Ho Kei Cheng and Piotr Dollár and Nikhila Ravi and Kate Saenko and Pengchuan Zhang and Christoph Feichtenhofer},
      year={2026},
      eprint={2511.16719},
      archivePrefix={arXiv},
      primaryClass={cs.CV},
      url={https://arxiv.org/abs/2511.16719}, 
}

@inproceedings{astruc2025anysat,
  title={Anysat: One earth observation model for many resolutions, scales, and modalities},
  author={Astruc, Guillaume and others},
  booktitle={CVPR},
  year={2025}
}

@inproceedings{weber2024flattening,
  title={Flattening the parent bias: Hierarchical semantic segmentation in the poincar{\'e} ball},
  author={Weber, Simon and Z{\"o}ng{\"u}r, Bar and Araslanov, Nikita and Cremers, Daniel},
  booktitle={Proceedings of the IEEE/CVF Conference on Computer Vision and Pattern Recognition},
  pages={28223--28232},
  year={2024}
}

@article{lin2022continual,
  title={Continual learning with evolving class ontologies},
  author={Lin, Zhiqiu and Pathak, Deepak and Wang, Yu-Xiong and Ramanan, Deva and Kong, Shu},
  journal={Advances in Neural Information Processing Systems},
  volume={35},
  pages={7671--7684},
  year={2022}
}

@article{weng2024mdinet,
  title={MDINet: Multidomain incremental network for change detection},
  author={Weng, Lean and Yang, Wenqing and Hu, Boni and Han, Pengcheng and Xue, Shaocheng and Zhang, Yu and Li, Haowei and Jin, Jie and Bu, Shuhui},
  journal={IEEE Transactions on Geoscience and Remote Sensing},
  volume={62},
  pages={1--15},
  year={2024},
  publisher={IEEE}
}

@article{tasar2019incremental,
  title={Incremental learning for semantic segmentation of large-scale remote sensing data},
  author={Tasar, Onur and Tarabalka, Yuliya and Alliez, Pierre},
  journal={IEEE Journal of Selected Topics in Applied Earth Observations and Remote Sensing},
  volume={12},
  number={9},
  pages={3524--3537},
  year={2019},
  publisher={IEEE}
}

@article{zhang2025foba,
  title={FoBa: A foreground--background co-guided method and new benchmark for remote sensing semantic change detection},
  author={Zhang, Haotian and Guo, Han and Chen, Keyan and Chen, Hao and Zou, Zhengxia and Shi, Zhenwei},
  journal={IEEE Transactions on Geoscience and Remote Sensing},
  volume={63},
  pages={1--19},
  year={2025},
  publisher={IEEE}
}

@article{zhang2023lim,
  title={LIM-CD: A large-scale remote sensing change detection dataset for incremental monitoring},
  author={Zhang, Hanchao and Zhang, Ruiqian and Ning, Xiaogang and Huang, Xiao and He, You and Chen, Yixin and Li, Mingzhu and Cui, Wei and Wang, Jiaming},
  journal={ISPRS Annals of the Photogrammetry, Remote Sensing and Spatial Information Sciences},
  volume={10},
  pages={903--910},
  year={2023},
  publisher={Copernicus GmbH}
}

@article{ghadimi2021hyperbolic,
  title={Hyperbolic busemann learning with ideal prototypes},
  author={Ghadimi Atigh, Mina and Keller-Ressel, Martin and Mettes, Pascal},
  journal={Advances in neural information processing systems},
  volume={34},
  pages={103--115},
  year={2021}
}

@article{garioud2025flair,
  title={FLAIR-HUB: Large-scale Multimodal Dataset for Land Cover and Crop Mapping},
  author={Garioud, Anatol and Giordano, S{\'e}bastien and David, Nicolas and Gonthier, Nicolas},
  journal={arXiv preprint arXiv:2506.07080},
  year={2025}
}

@article{tian2020hi,
  title={Hi-UCD: A large-scale dataset for urban semantic change detection in remote sensing imagery},
  author={Tian, Shiqi and Ma, Ailong and Zheng, Zhuo and Zhong, Yanfei},
  journal={arXiv preprint arXiv:2011.03247},
  year={2020}
}

@article{yang2020semantic,
  title={Semantic change detection with asymmetric Siamese networks},
  author={Yang, Kunping and Xia, Gui-Song and Liu, Zicheng and Du, Bo and Yang, Wen and Pelillo, Marcello and Zhang, Liangpei},
  journal={arXiv preprint arXiv:2010.05687},
  year={2020}
}

@article{zhou2023signet,
  title={Signet: A siamese graph convolutional network for multi-class urban change detection},
  author={Zhou, Yanpeng and Wang, Jinjie and Ding, Jianli and Liu, Bohua and Weng, Nan and Xiao, Hongzhi},
  journal={Remote Sensing},
  volume={15},
  number={9},
  pages={2464},
  year={2023},
  publisher={MDPI}
}

@inproceedings{toker2022dynamicearthnet,
  title={Dynamicearthnet: Daily multi-spectral satellite dataset for semantic change segmentation},
  author={Toker, Aysim and Kondmann, Lukas and Weber, Mark and Eisenberger, Marvin and Camero, Andr{\'e}s and Hu, Jingliang and Hoderlein, Ariadna Pregel and {\c{S}}enaras, {\c{C}}a{\u{g}}lar and Davis, Timothy and Cremers, Daniel and others},
  booktitle={Proceedings of the IEEE/CVF Conference on Computer Vision and Pattern Recognition},
  pages={21158--21167},
  year={2022}
}

@article{zheng2024segment,
  title={Segment any change},
  author={Zheng, Zhuo and Zhong, Yanfei and Zhang, Liangpei and Ermon, Stefano},
  journal={Advances in Neural Information Processing Systems},
  volume={37},
  pages={81204--81224},
  year={2024}
}

@article{bergamasco2022unsupervised,
  title={Unsupervised change detection using convolutional-autoencoder multiresolution features},
  author={Bergamasco, Luca and Saha, Sudipan and Bovolo, Francesca and Bruzzone, Lorenzo},
  journal={IEEE Transactions on Geoscience and Remote Sensing},
  volume={60},
  pages={1--19},
  year={2022},
  publisher={IEEE}
}

@inproceedings{sumbul2025smarties,
  title={SMARTIES: Spectrum-aware multi-sensor auto-encoder for remote sensing images},
  author={Sumbul, Gencer and Xu, Chang and Dalsasso, Emanuele and Tuia, Devis},
  booktitle={Proceedings of the IEEE/CVF International Conference on Computer Vision},
  pages={5569--5578},
  year={2025}
}

@article{qiu2023sats,
  title={SATS: Self-attention transfer for continual semantic segmentation},
  author={Qiu, Yiqiao and Shen, Yixing and Sun, Zhuohao and Zheng, Yanchong and Chang, Xiaobin and Zheng, Weishi and Wang, Ruixuan},
  journal={Pattern Recognition},
  volume={138},
  pages={109383},
  year={2023},
  publisher={Elsevier}
}

@article{cong2022satmae,
  title={Satmae: Pre-training transformers for temporal and multi-spectral satellite imagery},
  author={Cong, Yezhen and Khanna, Samar and Meng, Chenlin and Liu, Patrick and Rozi, Erik and He, Yutong and Burke, Marshall and Lobell, David and Ermon, Stefano},
  journal={Advances in Neural Information Processing Systems},
  volume={35},
  pages={197--211},
  year={2022}
}

@article{du2018unsupervised,
  title={Unsupervised deep slow feature analysis for change detection in multi-temporal remote sensing images},
  author={Du, Bo and Ru, Lixiang and Wu, Chen and Zhang, Liangpei},
  journal={arXiv preprint arXiv:1812.00645},
  year={2018}
}

@article{lei2026remote,
  title={Remote sensing image change detection using deep learning techniques: a comprehensive survey},
  author={Lei, Tao and Zhang, Shuxin and Lin, Shaoxiong and Liu, Tongfei and Lv, Zhiyong and Gao, Tao and Gong, Maoguo and Nandi, Asoke K},
  journal={Artificial Intelligence Review},
  year={2026},
  publisher={Springer}
}

@article{saha2019unsupervised,
  title={Unsupervised deep change vector analysis for multiple-change detection in VHR images},
  author={Saha, Sudipan and Bovolo, Francesca and Bruzzone, Lorenzo},
  journal={IEEE transactions on geoscience and remote sensing},
  volume={57},
  number={6},
  pages={3677--3693},
  year={2019},
  publisher={IEEE}
}

@inproceedings{goswami2023attribution,
  title={Attribution-aware weight transfer: A warm-start initialization for class-incremental semantic segmentation},
  author={Goswami, Dipam and Schuster, Ren{\'e} and Van de Weijer, Joost and Stricker, Didier},
  booktitle={Proceedings of the IEEE/CVF Winter Conference on Applications of Computer Vision},
  pages={3195--3204},
  year={2023}
}

@article{wu2021unsupervised,
  title={Unsupervised change detection in multitemporal VHR images based on deep kernel PCA convolutional mapping network},
  author={Wu, Chen and Chen, Hongruixuan and Du, Bo and Zhang, Liangpei},
  journal={IEEE Transactions on Cybernetics},
  volume={52},
  number={11},
  pages={12084--12098},
  year={2021},
  publisher={IEEE}
}

@article{kondmann2021spatial,
  title={Spatial context awareness for unsupervised change detection in optical satellite images},
  author={Kondmann, Lukas and Toker, Aysim and Saha, Sudipan and Sch{\"o}lkopf, Bernhard and Leal-Taix{\'e}, Laura and Zhu, Xiao Xiang},
  journal={IEEE Transactions on Geoscience and Remote Sensing},
  volume={60},
  pages={1--15},
  year={2021},
  publisher={IEEE}
}

@inproceedings{ji2024changenet,
  title={Changenet: Multi-temporal asymmetric change detection dataset},
  author={Ji, Deyi and Gao, Siqi and Tao, Mingyuan and Lu, Hongtao and Zhao, Feng},
  booktitle={ICASSP 2024-2024 IEEE International Conference on Acoustics, Speech and Signal Processing (ICASSP)},
  pages={2725--2729},
  year={2024},
  organization={IEEE}
}

@article{li2023semantic,
  title={Semantic hierarchy-aware segmentation},
  author={Li, Liulei and Wang, Wenguan and Zhou, Tianfei and Quan, Ruijie and Yang, Yi},
  journal={IEEE Transactions on Pattern Analysis and Machine Intelligence},
  volume={46},
  number={4},
  pages={2123--2138},
  year={2023},
  publisher={IEEE}
}

@article{katsmanRiemannianResidualNeural2023,
  title={Riemannian residual neural networks},
  author={Katsman, Isay and Chen, Eric and Holalkere, Sidhanth and Asch, Anna and Lou, Aaron and Lim, Ser Nam and De Sa, Christopher M},
  journal={Advances in Neural Information Processing Systems},
  volume={36},
  pages={63502--63514},
  year={2023}
}

@article{falorsiNeuralOrdinaryDifferential2020,
  title={Neural ordinary differential equations on manifolds},
  author={Falorsi, Luca and Forr{\'e}, Patrick},
  journal={arXiv preprint arXiv:2006.06663},
  year={2020}
}

@article{katsmanEquivariantManifoldFlows2021,
  title={Equivariant manifold flows},
  author={Katsman, Isay and Lou, Aaron and Lim, Derek and Jiang, Qingxuan and Lim, Ser Nam and De Sa, Christopher M},
  journal={Advances in Neural Information Processing Systems},
  volume={34},
  pages={10600--10612},
  year={2021}
}

@article{louNeuralManifoldOrdinary2020,
  title={Neural manifold ordinary differential equations},
  author={Lou, Aaron and Lim, Derek and Katsman, Isay and Huang, Leo and Jiang, Qingxuan and Lim, Ser Nam and De Sa, Christopher M},
  journal={Advances in Neural Information Processing Systems},
  volume={33},
  pages={17548--17558},
  year={2020}
}

@article{mathieuRiemannianContinuousNormalizing2020,
  title={Riemannian continuous normalizing flows},
  author={Mathieu, Emile and Nickel, Maximilian},
  journal={Advances in neural information processing systems},
  volume={33},
  pages={2503--2515},
  year={2020}
}

@inproceedings{vanspenglerPoincareResNet2023,
  title={Poincare resnet},
  author={van Spengler, Max and Berkhout, Erwin and Mettes, Pascal},
  booktitle={Proceedings of the IEEE/CVF International Conference on Computer Vision},
  pages={5419--5428},
  year={2023}
}

@inproceedings{rezendeNormalizingFlowsTori2020,
  title={Normalizing flows on tori and spheres},
  author={Rezende, Danilo Jimenez and Papamakarios, George and Racaniere, S{\'e}bastien and Albergo, Michael and Kanwar, Gurtej and Shanahan, Phiala and Cranmer, Kyle},
  booktitle={International Conference on Machine Learning},
  pages={8083--8092},
  year={2020},
  organization={PMLR}
}

@article{bergMultiPrototypeHyperbolicLearning2025,
  title={Multi-Prototype Hyperbolic Learning Guided by Class Hierarchy: P. Berg et al.},
  author={Berg, Paul and Buecher, L{\'e}o and Michele, Bj{\"o}rn and Pham, Minh-Tan and Chapel, Laetitia and Courty, Nicolas},
  journal={International Journal of Computer Vision},
  volume={133},
  number={11},
  pages={7969--7984},
  year={2025},
  publisher={Springer}
}

@inproceedings{chamiHoroPCAHyperbolicDimensionality2021,
  title = {{{HoroPCA}}: {{Hyperbolic Dimensionality Reduction}} via {{Horospherical Projections}}},
  shorttitle = {{{HoroPCA}}},
  booktitle = {Proceedings of the 38th {{International Conference}} on {{Machine Learning}}},
  author = {Chami, Ines and Gu, Albert and Nguyen, Dat P. and Re, Christopher},
  year = 2021,
  month = jul,
  pages = {1419--1429},
  publisher = {PMLR},
  issn = {2640-3498},
  urldate = {2024-05-27},
  langid = {english}
}

@article{fanHorosphericalDecisionBoundaries2023,
  title = {Horospherical {{Decision Boundaries}} for {{Large Margin Classification}} in {{Hyperbolic Space}}},
  shorttitle = {{{HoroSVM}}},
  author = {Fan, Xiran and Yang, Chun-Hao and Vemuri, Baba},
  year = 2023,
  month = dec,
  journal = {Advances in Neural Information Processing Systems},
  volume = {36},
  pages = {11194--11204},
  urldate = {2024-09-05},
  langid = {english}
}

@inproceedings{ganeaHyperbolicEntailmentCones2018,
  title = {Hyperbolic {{Entailment Cones}} for {{Learning Hierarchical Embeddings}}},
  booktitle = {Proceedings of the 35th {{International Conference}} on {{Machine Learning}}},
  author = {Ganea, Octavian and Becigneul, Gary and Hofmann, Thomas},
  year = 2018,
  month = jul,
  pages = {1646--1655},
  publisher = {PMLR},
  issn = {2640-3498},
  urldate = {2024-09-12},
  langid = {english}
}

@inproceedings{ganeaHyperbolicNeuralNetworks2018,
  title = {Hyperbolic {{Neural Networks}}},
  shorttitle = {{{HNN}}},
  booktitle = {Advances in {{Neural Information Processing Systems}}},
  author = {Ganea, Octavian and Becigneul, Gary and Hofmann, Thomas},
  year = 2018,
  volume = {31},
  publisher = {Curran Associates, Inc.},
  urldate = {2024-06-04}
}

@article{nickelPoincareEmbeddingsLearning2017,
  title={Poincar{\'e} embeddings for learning hierarchical representations},
  author={Nickel, Maximillian and Kiela, Douwe},
  journal={Advances in neural information processing systems},
  volume={30},
  year={2017}
}

@inproceedings{shimizuHyperbolicNeuralNetworks2020,
  title = {Hyperbolic {{Neural Networks}}++},
  shorttitle = {{{HNN}}++},
  booktitle = {International {{Conference}} on {{Learning Representations}}},
  author = {Shimizu, Ryohei and Mukuta, Yusuke and Harada, Tatsuya},
  year = 2020,
  month = oct,
  urldate = {2026-01-30},
  langid = {english}
}

@inproceedings{choLargeMarginClassificationHyperbolic2019,
  title = {Large-{{Margin Classification}} in {{Hyperbolic Space}}},
  shorttitle = {Hyperbolic {{SVM}}},
  booktitle = {Proceedings of the {{Twenty-Second International Conference}} on {{Artificial Intelligence}} and {{Statistics}}},
  author = {Cho, Hyunghoon and DeMeo, Benjamin and Peng, Jian and Berger, Bonnie},
  year = 2019,
  month = apr,
  pages = {1832--1840},
  publisher = {PMLR},
  issn = {2640-3498},
  urldate = {2024-09-12},
  langid = {english}
}

@inproceedings{malila1980change,
  title={Change vector analysis: An approach for detecting forest changes with Landsat},
  author={Malila, William A},
  booktitle={LARS symposia},
  pages={385},
  year={1980}
}

@article{nielsen1997multivariate,
  title={Multivariate alteration detection (MAD) in multispectral, bi-temporal image data: a new approach to change detection studies},
  author={Nielsen, Allan Aasbjerg and Conradsen, Knut},
  journal={Dept. Mathematical Modelling, Tech. Univ. Denmark, Lyngby, Denmark, Tech. Rep},
  volume={11},
  pages={1997},
  year={1997}
}

@article{volpi2012unsupervised,
  title={Unsupervised change detection with kernels},
  author={Volpi, Michele and Tuia, Devis and Camps-Valls, Gustavo and Kanevski, Mikhail},
  journal={IEEE Geoscience and Remote Sensing Letters},
  volume={9},
  number={6},
  pages={1026--1030},
  year={2012},
  publisher={IEEE}
}

@article{volpi2013supervised,
  title={Supervised change detection in VHR images using contextual information and support vector machines},
  author={Volpi, Michele and Tuia, Devis and Bovolo, Francesca and Kanevski, Mikhail and Bruzzone, Lorenzo},
  journal={International Journal of Applied Earth Observation and Geoinformation},
  volume={20},
  pages={77--85},
  year={2013},
  publisher={Elsevier}
}

@inproceedings{douillard2021plop,
  title={Plop: Learning without forgetting for continual semantic segmentation},
  author={Douillard, Arthur and Chen, Yifu and Dapogny, Arnaud and Cord, Matthieu},
  booktitle={Proceedings of the IEEE/CVF conference on computer vision and pattern recognition},
  pages={4040--4050},
  year={2021}
}

@book{bridsonMetricSpacesNonPositive1999,
  title = {Metric {{Spaces}} of {{Non-Positive Curvature}}},
  author = {Bridson, Martin R. and Haefliger, Andr{\'e}},
  year = 1999,
  series = {Grundlehren Der Mathematischen {{Wissenschaften}}},
  volume = {319},
  publisher = {Springer},
  address = {Berlin, Heidelberg},
  issn = {0072-7830},
  
  urldate = {2025-07-11},
  copyright = {https://www.springernature.com/gp/researchers/text-and-data-mining},
  isbn = {978-3-642-08399-0 978-3-662-12494-9},
  langid = {english}
}

@article{bonnabelStochasticGradientDescent2013,
  title = {Stochastic {{Gradient Descent}} on {{Riemannian Manifolds}}},
  author = {Bonnabel, Silv{\`e}re},
  year = 2013,
  month = sep,
  journal = {IEEE Transactions on Automatic Control},
  volume = {58},
  number = {9},
  pages = {2217--2229},
  issn = {1558-2523},
  doi = {10.1109/TAC.2013.2254619},
  urldate = {2025-01-06}
}
\bibliographystyle{iclr2027_conference}

\newpage
\appendix
\section*{Appendix}
\paragraph{Organization of the Appendix.} These supplementary materials provide additional details and experiments complementing the main paper: \begin{itemize}
    \item \Cref{sec:supp_hichange} presents further information about the proposed HiChange dataset, including comparisons with existing SCD datasets, the sampling strategy, the label generation pipeline, and dataset statistics.
    \item \Cref{sec:supp_other_ds} summarizes the characteristics of the SECOND and Hi-UCD-mini datasets used in our experiments.
    \item \Cref{sec:supp_hyperbolics} reviews the hyperbolic geometry concepts underlying our method.  \item \Cref{sec:supp_implementation} provides implementation details and the definition of the incremental tasks.
    \item \Cref{sec:supp_qualitative} presents additional qualitative results.
    \item \Cref{sec:supp_experiments} reports complementary experiments analyzing the influence of the threshold, backbone architecture and embedding dimension.
\end{itemize}

\section{Additional Information on the HiChange Dataset}
\label{sec:supp_hichange}

\subsection{Comparison with Existing SCD Datasets}

\Cref{tab:datasets} compares HiChange with commonly used SCD datasets. The latter are typically limited to a small number of semantic categories. This limitation motivates the creation of HiChange, as such restricted semantic diversity makes them less suitable for studying CI-SCD scenarios. HiChange addresses these limitations by introducing a larger and hierarchically structured taxonomy designed to support progressive semantic refinement. 

\begin{table}[!h]
\centering
\footnotesize
\caption{Existing SCD datasets, and comparison with the proposed HiChange dataset.}
\label{tab:datasets}
\begin{tabular}{@{}ccccc@{}}
\toprule
\textbf{Dataset} & \textbf{Image pairs}                                                    & \textbf{Image size}    & \textbf{Res {[}m{]}} & \textbf{Categories}                                           \\ \midrule
HRSCD \cite{daudt2019multitask}           & 291                                                                     & $10000\times10000$     & 0.5                         & 5                                                                      \\
Hi-UCD-mini \cite{tian2020hi}          & 1293                                                                    & $1024\times1024$       & 0.1                         & 9                                                                      \\
SECOND \cite{yang2020semantic}          & 4662                                                                    & $512\times512$         & not indicated               & 6                                                                      \\
CNAM-CD \cite{zhou2023signet}         & 2503                                                                    & $512\times512$         & 0.5                         & 5                                                                      \\
DynamicEarthNet \cite{toker2022dynamicearthnet} & \begin{tabular}[c]{@{}c@{}}75 time series \\ of 730 images\end{tabular} & $1024\times1024$       & 3                           & 7                                                                      \\
ChangeNet \cite{ji2024changenet}        & 31000                                                                   & $\sim 1900\times 1200$ & 0.3                         & 6                                                                      \\
LevirSCD \cite{zhang2025foba}        & 3225                                                                   & $256\times 256$ & 1-2                         & 16                                                                      \\\midrule
HiChange         & 1000                                                                    & $5000\times5000$         & 0.2                         & \begin{tabular}[c]{@{}c@{}}4 - 32, organized \\ hierarchically\end{tabular} \\ \bottomrule
\end{tabular}
\end{table}

\subsection{Sampling of the Images}

In this section, we detail the sampling procedure used to construct the HiChange dataset. The objective was to select a subset of tiles from the French national BDOrtho aerial imagery database\footnote{\url{https://geoservices.ign.fr/bdortho}} that maximizes the diversity and relevance of observed semantic changes. We fixed the total number of tiles to $N = 1000$, a choice made to balance dataset size, semantic diversity, and feasible training time for deep-learning models. Each tile has a spatial resolution of $0.2$ m and a size of $5000\times 5000$ pixels. The sampling procedure consists of three main stages described below.

\begin{figure}
    \centering
    \includegraphics[width=\linewidth]{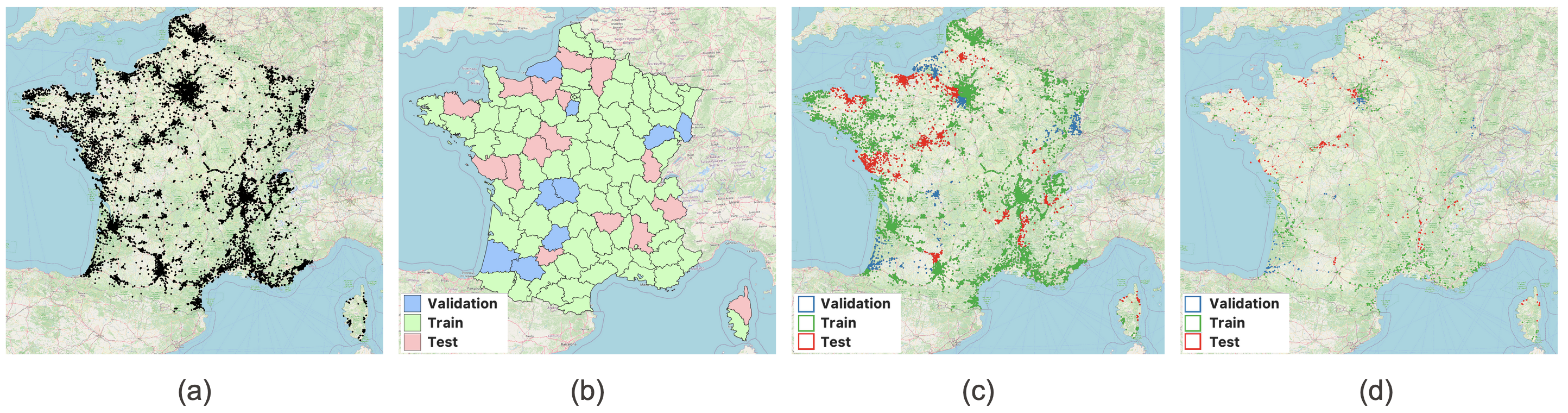}
    \caption{Sampling procedure for constructing the HiChange dataset. (a) Tiles retained after the initial filtering step based on the number of changes observed. (b) Geographical partitioning of French administrative departments used to define train, test and validation splits. (c) Assignment of tiles to the corresponding split according to their geographical location. (d) Final selection of the tiles after the greedy sampling algorithm, ensuring diverse and balanced semantic change types in all splits of the dataset.}
    \label{fig:sampling_procedure}
\end{figure}

\paragraph{Initial Filtering Based on the Number of Changes Observed.} To ensure that the final dataset focuses on regions with actual changes, we perform an initial selection of tiles based on the density of changes within each tile. For this purpose, we use the OCSGE database\footnote{\url{https://geoservices.ign.fr/ocsge}}, which we also employ for change label generation. As OCSGE is available only for specific years depending on the region, we select BDOrtho acquisition years to match the available OCSGE years to ensure temporal consistency. Acquisition years range from 2016 to 2023, and time between the 2 acquisitions for each pair is 2 to 3 years. OCSGE is provided as a vector dataset in which each changed object is represented as a polygon. For each tile in BDOrtho, we count the number of OCSGE polygons falling within its spatial extent. We retain only tiles containing more than $30$ polygons of change. This threshold was chosen to filter out tiles without meaningful change activity. This procedure yields approximately $20000$ tiles. As anticipated, the retained tiles concentrate around dynamic, predominantly urban, regions as depicted in \cref{fig:sampling_procedure}, panel (a). 

\paragraph{Geographical train-validation-test Assignment.} The second step consists in defining the dataset splits. We opt for a geographical partitioning based on administrative departments to prevent spatial overlap between training, validation, and testing. Each of the roughly $20000$ candidate tiles is assigned to one of the three splits (train/validation/test), as depicted in \cref{fig:sampling_procedure}, panel (b) and (c).

\paragraph{Greedy Selection for Diversity and Class Balance.} Even after the initial filtering, the number of candidate tiles remains far larger than the desired dataset size. The final selection aims to (i) maximize diversity in the types of observed semantic changes and (ii) balance the relative frequency of change categories. To this end, we exploit the land-cover attributes provided by OCSGE and adopt an iterative greedy selection procedure. For each change polygon, its change type is defined as the combination of the dominant semantic labels before and after change (\ie, an ordered pair $(\mathrm{category}_\mathrm{before},\mathrm{category}_\mathrm{after})$). For each train-test-validation split, we run the tile selection procedure as follows: \begin{enumerate}
    \item Initialize the selected tile set as empty.
    \item For each unselected tile in the split: \begin{itemize}
        \item Simulate adding the tile to the current selection.
        \item Compute the deviation between the resulting distribution of change types and the target uniform distribution across all change types.
        \item Measure the reduction in deviation that would result from adding the tile.
    \end{itemize}
    \item Select the tile that yields the greatest improvement.
    \item Repeat until the desired number of tiles $K$ for the split is reached.
\end{enumerate}

This greedy algorithm progressively builds a tile set that is both diverse and balanced in terms of semantic change types. We sample $K=700$ tiles for the train set, $K=100$ for the validation set and $K=200$ for the test set. The final selected tiles are depicted in \cref{fig:sampling_procedure}, panel (d).

\subsection{Generation of the Labels}

\begin{figure}
    \centering
    \includegraphics[width=\linewidth]{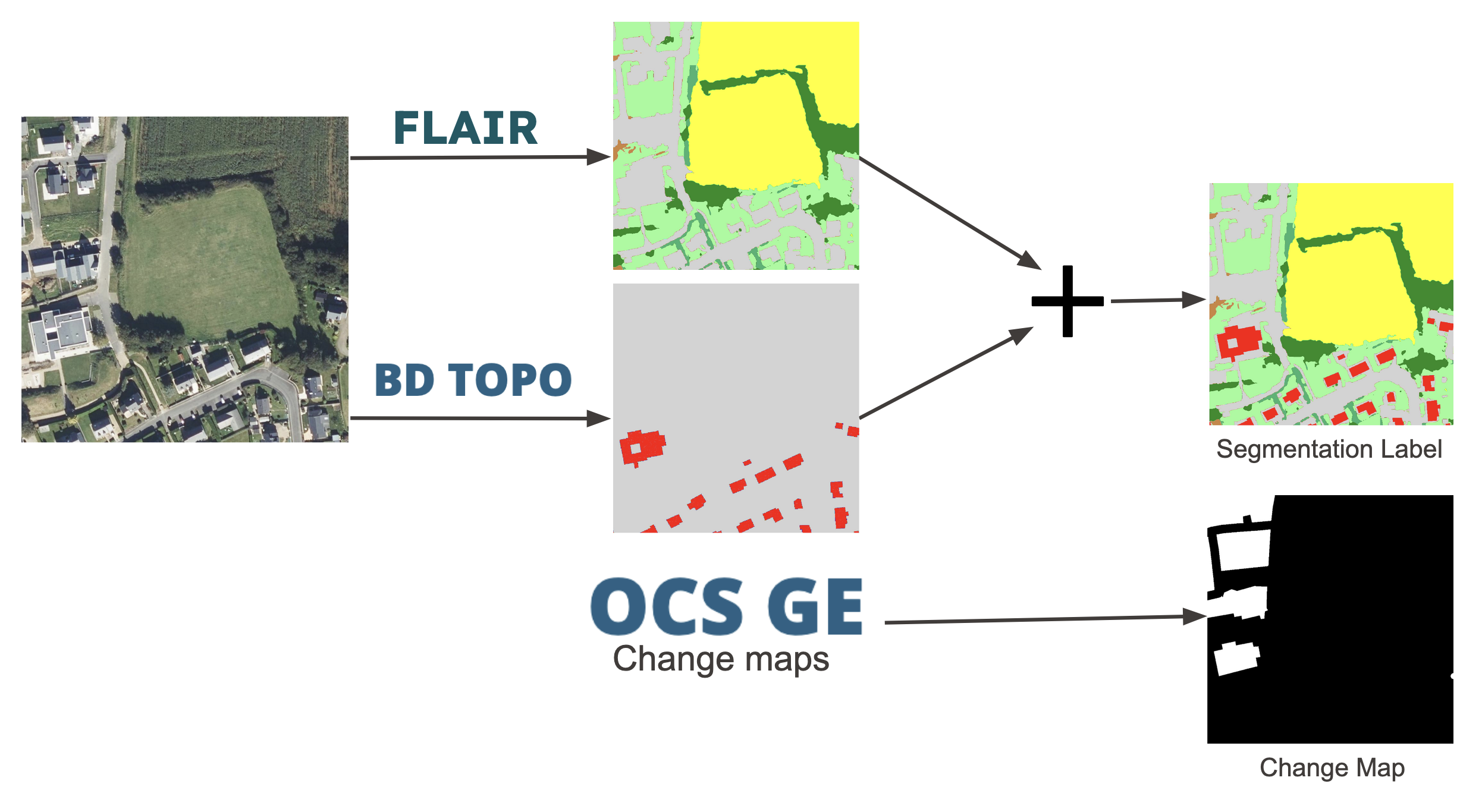}
    \caption{Pipeline for the generation of the labels. }
    \label{fig:pipeline_generation_labels}
\end{figure}

The pipeline for the generation of the labels is presented in Figure \ref{fig:pipeline_generation_labels}. The change detection labels are derived from the OCSGE database, which provides coarse annotations of both land-cover segmentation and change between two temporal acquisitions. We directly exploit the change information from OCSGE as a binary indication of change. This change map is later refined at usage time to align with the selected semantic granularity, as detailed later in this section. However, the semantic segmentation labels provided by OCSGE remain coarse and cover only a limited set of semantic categories\footnote{\url{https://geoservices.ign.fr/sites/default/files/2025-10/DC_OCSGE_NG.pdf}}. This level of detail is insufficient for the objectives of HiChange, which require fine-grained semantic information. Therefore, additional sources of information are incorporated to construct richer semantic labels. In particular, we leverage the BDTopo database\footnote{\url{https://geoservices.ign.fr/bdtopo}} from the French National Institute of Geographic and Forest Information (IGN) which provides high-precision vector layers describing buildings, roads, and other man-made infrastructure for multiple years. A key limitation, however, is that BDTopo is not spatially exhaustive: several regions of the territory remain unlabeled or only partially covered. One possible remedy would be to supplement BDTopo with the OCSGE semantic segmentation data, but the latter is too coarse for our target spatial resolution ($0.2$ m). To overcome this limitation, we employ the FLAIR land-cover model \cite{garioud2025flair}, which was also trained on BDOrtho imagery and has demonstrated high accuracy on segmentation tasks. We run FLAIR inference over all tiles and use its predicted land-cover labels in regions not covered by BDTopo. The resulting semantic segmentation combines BDTopo-derived vector information where available and FLAIR predictions elsewhere, thereby achieving both fine-grained semantic resolution and full spatial coverage.

\begin{figure}
    \centering
    \includegraphics[width=\linewidth]{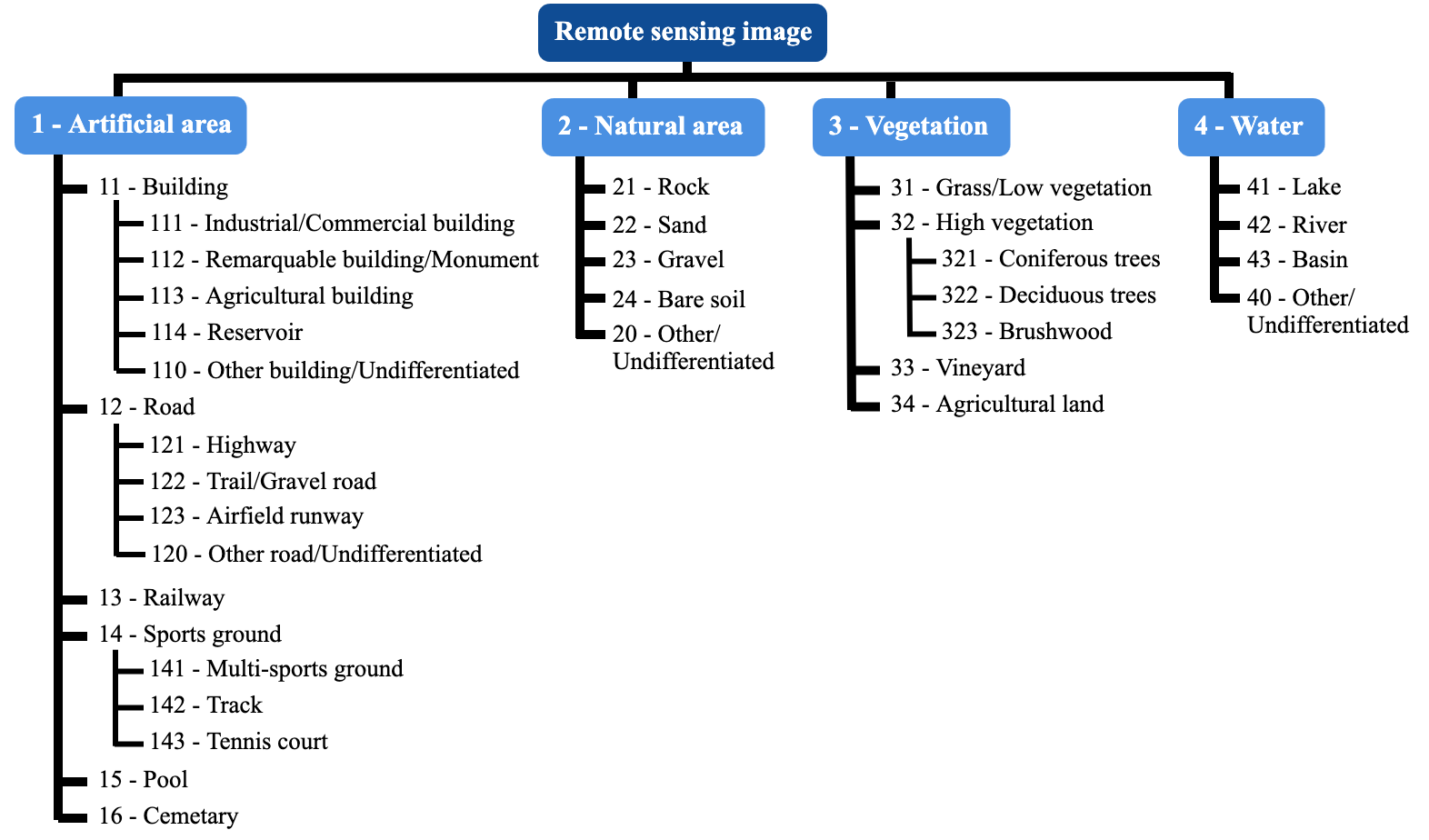}
    \caption{Taxonomy of the semantic labels of the HiChange dataset.}
    \label{fig:hierarchy_labels}
\end{figure}

\paragraph{Hierarchical Label Selection.} The above procedure yields semantic labels at the finest level of granularity. These labels are organized into a hierarchical taxonomy structured in three levels, as shown in \cref{fig:hierarchy_labels}. When using the dataset, the user selects the desired classes to consider. Fine-grained categories are then automatically mapped to their parent categories when coarser levels are requested.

\paragraph{Consistency Filtering of Change Labels.} To ensure coherence between the semantic segmentation labels and the change annotations, we apply a filtering mechanism to the change map at usage time. Specifically, if the selected classes are coarse and a tile exhibits a change only between child categories belonging to the same parent category (\eg, a change from \textit{forest} to \textit{grass} when only \textit{vegetation} is considered), the change should not be reported. We therefore mark as changed only pixels that are both (i) marked as changed by OCSGE and (ii) present different semantic labels between the two timestamps at the chosen semantic level. This guarantees that the binary change masks remain semantically meaningful and consistent across hierarchical configurations, and solves at the same time the problem of the coarse resolution of OCSGE.

\paragraph{Limitations} The construction of such a large dataset involves a trade-off between annotation quality and coverage. Some regions in the dataset where BDTopo is unavailable are labeled using FLAIR. These automatically generated labels may inherit biases and errors from the model, and may therefore introduce some noise into the benchmark. In addition, the use of OCSGE may exclude changes that fall below its minimum mapping unit. However, given the scale of the dataset, manual annotation is not practically feasible. The adopted methodology provides a compromise that enables large-scale evaluation while acknowledging the limitations associated with automatically generated labels.

\subsection{Examples from the dataset}

\begin{figure}[h!]
    \centering
    \includegraphics[width=\linewidth]{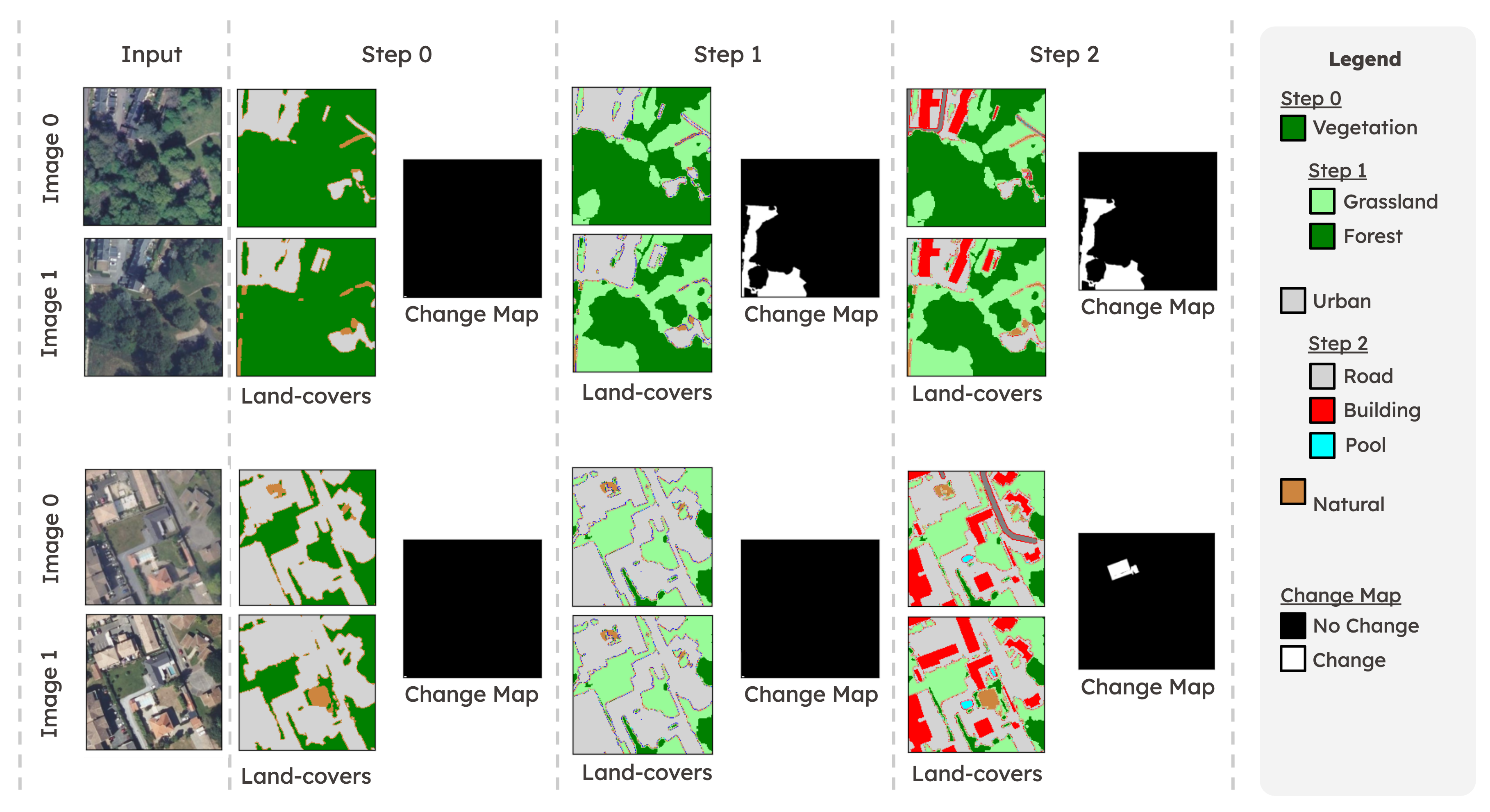}
    \caption{Examples from the HiChange dataset, illustrating the refinement of semantic labels. In the initial step, only coarse distinctions between vegetation and constructed areas are considered. At step 1, we have additional information on the type of vegetation and, at step 2, about the type of construction. Note that the change maps adapt to the set of semantic labels considered: in the first example, change happens within the vegetation, and is captured only at steps 1 and 2. Similarly, in the second example, a building is constructed and the change is only captured in the last incremental step.}
    \label{fig:HiChange_example}
\end{figure}

Some example images from the HiChange dataset are provided in \cref{fig:HiChange_example}.

\begin{figure}[h!]
  \centering
\begin{subfigure}{\textwidth}
  \centering
  \includegraphics[width=\linewidth]{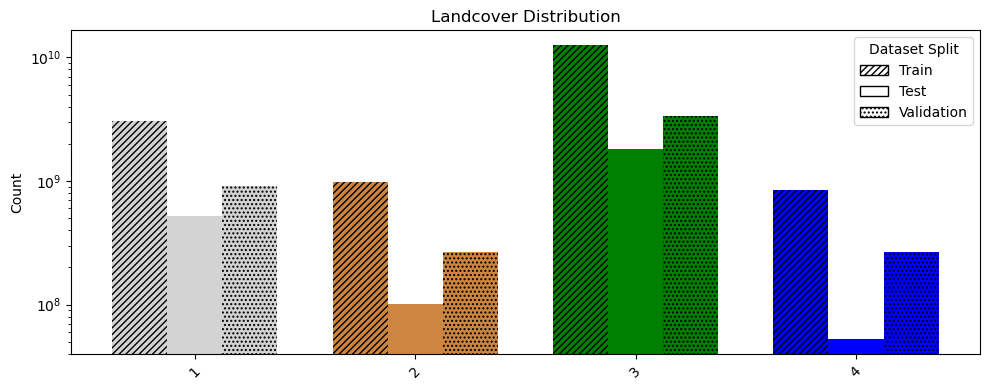}  
  \caption{}
\end{subfigure}\\
\begin{subfigure}{\textwidth}
  \centering
  \includegraphics[width=\linewidth]{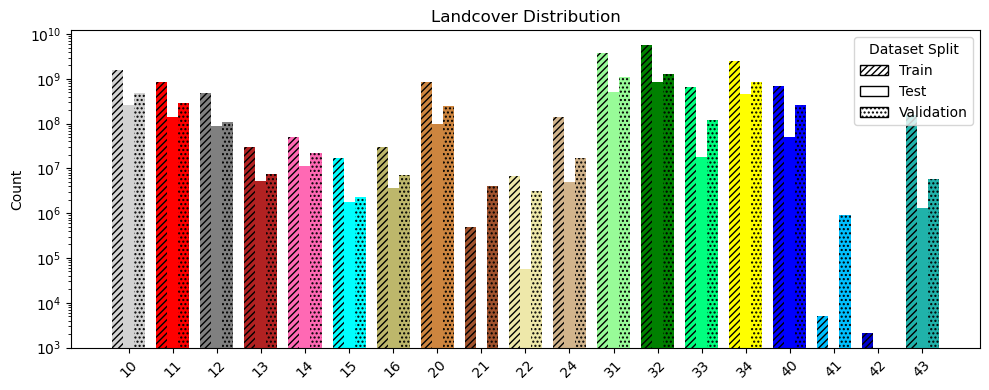}  
  \caption{}
\end{subfigure}\\
\begin{subfigure}{\textwidth}
  \centering
  \includegraphics[width=\linewidth]{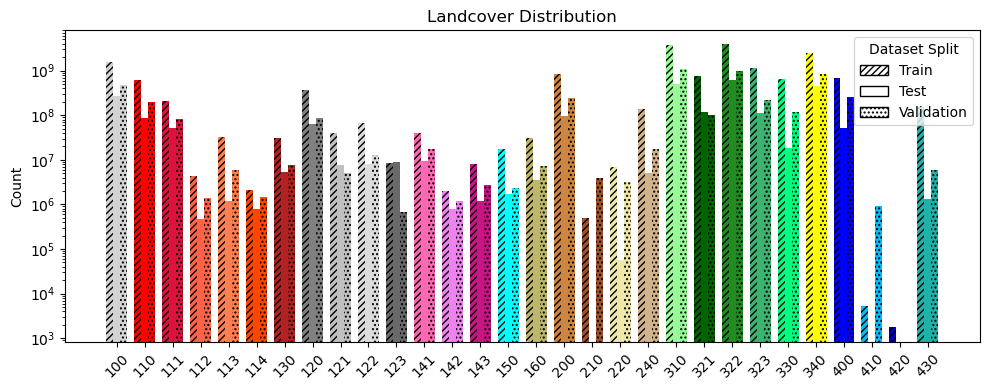}  
  \caption{}
\end{subfigure}
\caption{Distribution of the land-cover categories in the HiChange dataset at different levels of the taxonomy.}
\label{fig:dataset_statistics}
\end{figure}

\begin{figure}[h!]
  \centering
\begin{subfigure}{\textwidth}
  \centering
  \includegraphics[width=\linewidth]{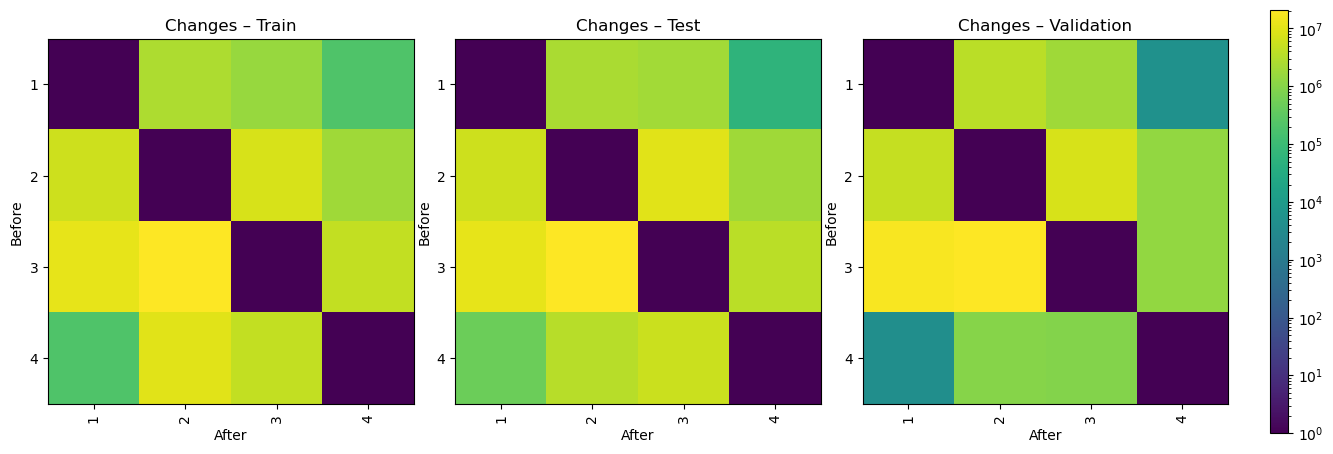}  
  \caption{}
\end{subfigure}
\begin{subfigure}{\textwidth}
  \centering
  \includegraphics[width=\linewidth]{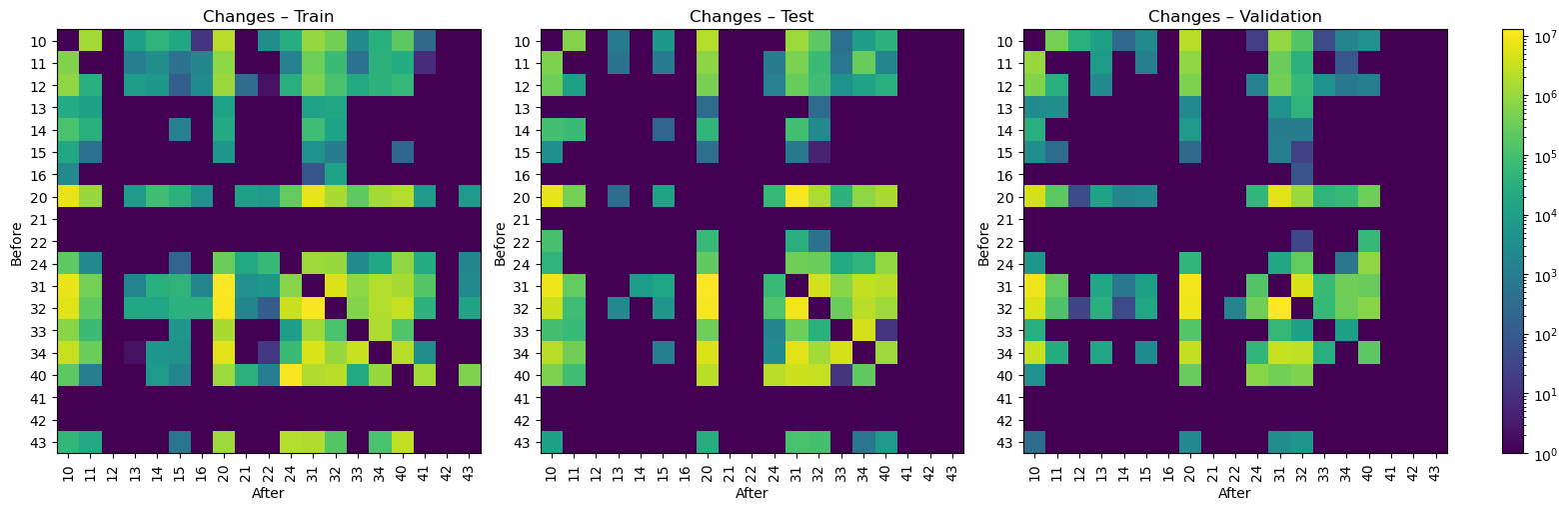}  
  \caption{}
\end{subfigure}
\begin{subfigure}{\textwidth}
  \centering
  \includegraphics[width=\linewidth]{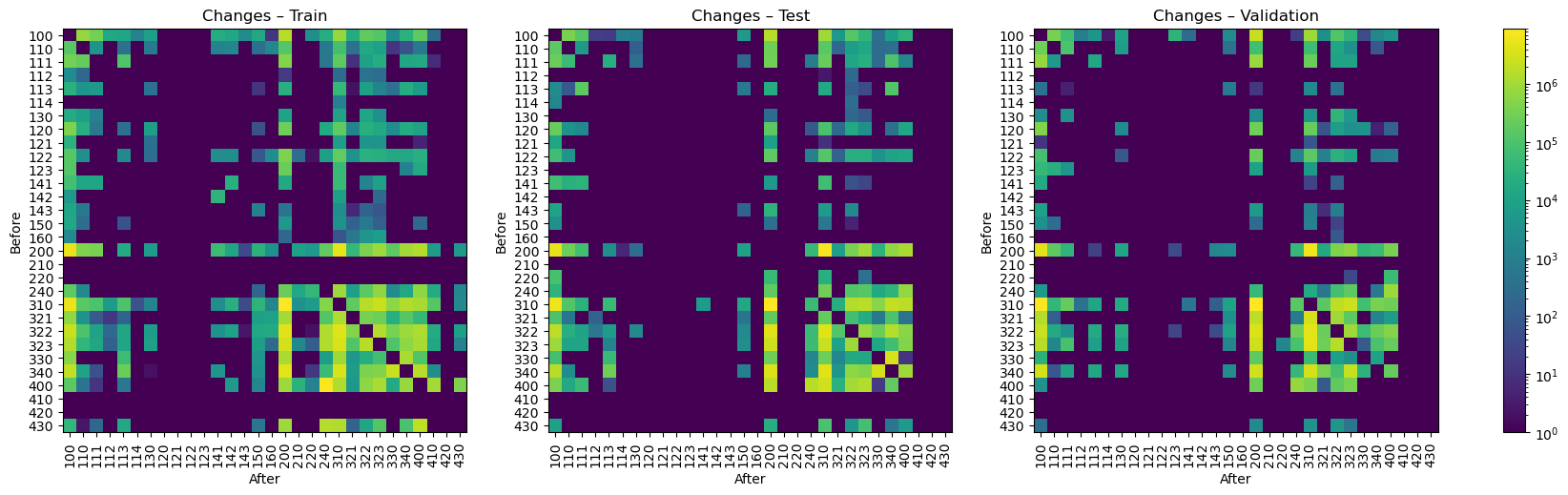}  
  \caption{}
\end{subfigure}
\caption{Semantic changes in HiChange. Confusion matrices of land-cover transitions for pixels labeled as change at the three hierarchical levels of the taxonomy.}
\label{fig:confusion_matrices_ds_stats}
\end{figure}

\subsection{Dataset Statistics}

The distribution of the semantic labels at the three levels of depth of the hierarchy are presented in \cref{fig:dataset_statistics}. The confusion matrix for the land-cover for pixels that contain change is presented in \cref{fig:confusion_matrices_ds_stats}. Although not all types of changes are present in the dataset, especially at the finest level of the hierarchy, we observe a large variety.

\section{Additional Information on the SECOND and Hi-UCD-mini Datasets}
\label{sec:supp_other_ds}

\paragraph{SECOND} The SECOND dataset consists of $4662$ pairs of $512\times 512$ aerial images collected from multiple platforms and sensors over several Chinese cities. Spatial resolution is not indicated. Images are annotated by experts on six land-cover classes: \textit{low vegetation, non-vegetated ground surface, trees, water, building} and \textit{playground}. Semantic annotation is provided for changed regions only, and therefore only evaluated on those regions. The dataset is split in $2968$ pairs for training and $1694$ for testing. 

\paragraph{Hi-UCD-mini} The Hi-UCD-mini dataset is an urban SCD dataset built from very-high-resolution aerial images ($0.1$ m spatial resolution). It focuses on the city of Tallinn and contains $1293$ image pairs of size $1024\times 1024$ acquired between 2017 and 2019. The images are annotated for both binary change and segmentation, covering $9$ semantic categories: \textit{water, grass, building, green house, road, bridge, bare land, woodland} and \textit{undefined}.

\section{Hyperbolic Space and Vector Fields}
\label{sec:supp_hyperbolics}

This section briefly reviews the hyperbolic geometry and vector field formulation that form the mathematical foundation of our method.

\subsection{Hyperbolic Space}
\label{sec:supp_hyperbolic}
Hyperbolic spaces are smooth manifolds, meaning that they can be endowed with Cartesian coordinates locally around every point. This structure makes it possible to define tangent vectors in each point, corresponding to the derivatives of all possible smooth curves going through it. The set of tangent vectors in some point $x$ of a manifold $\mathcal{M}$ is called the tangent space at $x$ and is denoted $T_x\mathcal{M}$. It is a vector space that can be seen as the first order approximation of the manifold around this point. 
Riemannian manifolds are smooth manifolds where each tangent space $T_x\mathcal{M}$ is endowed with a scalar product $g_x: T_x\mathcal{M}\times T_x\mathcal{M} \to \mathbb{R}$, called the Riemannian metric, and which completely characterizes the geometric structure of $\mathcal{M}$. 
Indeed, it makes it possible to define the length of a smooth curve $\gamma: [a, b]\to \mathcal{M}$ as 
\begin{equation}
    L(\gamma) = \int_a^b \sqrt{g_{\gamma(t)}(\dot \gamma(t), \dot\gamma(t))}\ \mathrm{d}t,
\end{equation}
and thus geodesics as minimal-length paths between two points. We can then define the distance between two points as the length of such a geodesic. By extension, a curve $\gamma:\mathbb{R}\to\mathcal{M}$ which is a geodesic on every segment $[a, b]$ is also called a geodesic.

Hyperbolic spaces are Riemannian manifolds with the particular property of having constant negative curvature. This curvature controls how much the space looks like a tree. The more negative, the more the geometry becomes comparable to that of an (infinite) tree. The closer to 0, the more the hyperbolic space gets similar to a Euclidean space. 
There exist different models of hyperbolic space that are, for a given dimension $d$ and curvature $\kappa < 0$, all isometric to each other. The Poincaré ball model is one of them. Using $c=-\kappa$, it is defined as
\begin{equation}
    \mathbb{B}_d = \left\{z\in\mathbb{R}^d \ | \ \|z\|^2_2 < \frac{1}{c} \right\},
\end{equation}
endowed with the Riemannian metric $g_x(u, v) = \lambda_x^2\langle u, v\rangle$, where $\lambda_x = \frac{2}{1-c\|x\|_2^2}$ is called the conformal factor.

In such a space, geodesics are given by arcs of circle (or straight lines) orthogonal to the "boundary" $\mathbb{S}^{d-1}_c = \{z\in\mathbb{R}^d \ | \ \|z\|_2^2 = 1/c \}$ (which is not reachable in finite time), and the Riemannian distance between two points $x, y\in\mathbb{B}_d$ is given by 
\begin{equation}
    d(x, y) = \frac{1}{\sqrt{c}}\cosh^{-1}\left(1 + 2c\frac{\|x-y\|_2^2}{(1-c\|x\|_2^2)(1-c\|y\|_2^2)}\right).
\end{equation}

Given a tangent vector $v \in T_x\mathbb{B}_d$ in a point $x\in\mathbb{B}_d$, the exponential map $\exp_x:T_x\mathbb{B}_d\to\mathbb{B}_d$ returns $\gamma_v(1)\in \mathbb{B}_d$ where $\gamma_v:\mathbb{R}\to\mathbb{B}_d$ is the constant-speed geodesic passing through $x$ in $t = 0$ with velocity $\gamma'(0) = v$, as such geodesic is unique. The exponential map therefore gives a way to move in the manifold $\mathbb{B}_d$ from the starting point $x$ following a displacement vector $v$. Its expression is given by
\begin{equation}
    \exp_x(v) = x \oplus \frac{1}{\sqrt{c}}\tanh\Big(\sqrt{c}\|v\|_2\Big)\frac{v}{\|v\|_2},
\end{equation}
where $\oplus$ denotes the Möbius addition, expressed, for $x, y\in\mathbb{B}_d$, as
\begin{equation}
    x \oplus y = \frac{(1 + 2c\langle x, y\rangle + c\|y\|_2^2)x + (1-c\|x\|_2^2) y}{1 + 2c\langle x, y\rangle + c^2\|x\|_2^2\|y\|_2^2}.
\end{equation}

\subsection{Busemann Function}

In a complete Riemannian manifold $\mathcal{M}$, for a unit-speed geodesic $\gamma:\mathbb{R}\to\mathcal{M}$, the Busemann function along $\gamma$ is defined as 
\begin{equation}
    B_\gamma: x\in \mathcal{M}\mapsto \lim_{t\to\infty} d(\gamma(t), x) - t, 
\end{equation}
when the limit exists. It can be interpreted as a sort of coordinate of the point $x$ along $\gamma$. 

This Busemann function is well-defined in Euclidean space, and for a unit-speed geodesic $\gamma_w : t\mapsto tw$, with $w\in\mathbb{S}^{d-1}$, going through the origin, we have $B_w(x) = -\langle x, w \rangle$. The Busemann function can therefore be seen as a generalization of the (negative) scalar product to other manifolds, and thus can be used as a tool for Riemannian machine learning.

In hyperbolic space, the Busemann function exists as well. For a prototype $p$ at infinity ($p\in\mathbb{S}^{d-1}_c$), which can be seen as a direction, the unit-speed geodesic \textit{going to} $p$ is written: 
\begin{equation}
    \gamma_p(t) = \exp_0\left(\frac{\sqrt{c}t}{2} p\right) =  \tanh\left(\frac{\sqrt{c}t}{2}\right)p.
\end{equation}
The Busemann function can then be shown to be
\begin{equation}
    B_p(x) = \frac{1}{\sqrt{c}}\log\left(\frac{c\|x - p\|_2^2}{1 - c\|x\|_2^2}\right).
\end{equation} 
Its level-sets
\begin{equation}
    H_{p, a} = \{z\in\mathbb{B}_d \ | \ -B_p(z) + a = 0\}
\end{equation}
are called horospheres. In the Poincaré model, they are represented by spheres tangent to the prototype. They can be seen as hyperbolic equivalents for Euclidean hyperplanes. Other generalizations of hyperplanes in hyperbolic space exist, such as geodesic hyperplanes that have been used in different machine learning works on hyperbolic geometry \cite{ganeaHyperbolicNeuralNetworks2018, choLargeMarginClassificationHyperbolic2019, shimizuHyperbolicNeuralNetworks2020}. However, one advantage of horospherical learning is that the distance from a point to a horosphere admits the simple expression
\begin{equation}
    d(x, H_{p, a}) = | B_p(x) - a |,
\end{equation}
which is analogous to the relation between Euclidean hyperplanes and scalar product. This expression is straightforward to compute and shows that the distance between two horospheres sharing the same prototype $p$ is constant. In particular, this distance is independent of the points at which it is evaluated, which provides a natural interpretation of horospherical classification.

\subsection{Busemann Vector Fields}
\label{sec:appendix-bvf}

The change detection part of HyperFlow rely on the flow of a hyperbolic vector field. In a general manifold $\mathcal{M}$, several possible approaches exist to define and parameterize such a vector field. A straightforward approach 
is to see the manifold $\mathcal{M}$ of dimension $d$ as a subset of a Euclidean embedding space $\mathbb{R}^n$ where $n\geq d$, and to naively project the vectors to their corresponding tangent space. However, this construction is  not geometrically grounded and may prove inefficient. 

We instead adopt the intrinsic approach proposed in \cite{katsmanRiemannianResidualNeural2023}. Let us assume there exists a parameterizable feature function $g_\theta:\mathcal{M}\to\mathbb{R}$ on the manifold. By considering $K$ different parameter values $\theta_1, ..., \theta_K$, we can construct a map  $f_\Theta:\mathcal{M}\to \mathbb{R}^K$ by concatenating the corresponding feature functions:
\begin{equation}
    f_\Theta = [g_{\theta_1}, ..., g_{\theta_K}],
\end{equation}
with $\Theta = (\theta_1, \dots, \theta_K)$. We interpret this function as a coordinate system on $\mathcal{M}$, although it is not required to be either injective or surjective. At a point $x\in\mathcal{M}$, its differential $D_xf_{\Theta}$ maps $T_x\mathcal{M}$ to $T_{f_\Theta(x)}\mathbb{R}^K \cong \mathbb{R}^K$. The corresponding pullback operator $(D_xf_{\Theta})^*$ maps the dual space $(\mathbb R^K)^*$ of $\mathbb{R}^K$ to the dual space $(T_x\mathcal M)^*$ of $T_x\mathcal M$. By identifying Euclidean spaces with their dual spaces, we can see the pullback operator as a map
\begin{equation}
    (D_xf_\Theta)^*: \mathbb R^K \to T_x\mathcal{M}.
\end{equation}
It is expressed by the gradients of the different coordinate features, which together form what we can see as a set of elementary vector fields.

Following \cite{katsmanRiemannianResidualNeural2023}, a vector field can then be defined on $\mathcal{M}$ as 
\begin{equation}
\label{eq:BVF}
    V_{\Theta, \omega}: x \mapsto (D_xf_{\Theta})^*\Big(n_\omega(f_\Theta(x))\Big) = \sum_{k=1}^K n_\omega(f_\Theta(x))_k\nabla g_{\theta_k}(x),
\end{equation}
where $n_\omega:\mathbb{R}^K\to\mathbb{R}^K$ is any neural network and $\nabla$ is the Riemannian gradient. While $f_\Theta$ feeds the neural network with some kind of coordinates, the outputs of this network serve as weights in a linear combination of the elementary vector fields. In the hyperbolic case, Busemann functions are natural candidates for feature functions, as proposed in \cite{katsmanRiemannianResidualNeural2023}.

\paragraph{Adaptation to Computer Vision Tasks.} Under the above formulation, the vector field, when applied to the embeddings of the pixels of an image, does not take the spatial context into account. Consequently, the displacement of a pixel embedding is determined solely by its position in space, independently of the embeddings of neighboring pixels. In this work, since we argue that the spatial context provides valuable information, we instead replace the neural network $n_\omega$, acting pixel-wise, by a neural network $N_\omega:(\mathbb{R}^K)^{H\times W}\to(\mathbb{R}^K)^{H\times W}$ that takes the full image of Busemann features as input (in our case, a UNet). The Busemann functions $f_P$ and their gradients $\nabla B_{p_k}$ are however still computed pixel-wise. This creates a vector field on $(\mathbb{B}_d)^{H\times W}$, and is equivalent to applying the design defined by \cref{eq:BVF} to $(\mathbb{B}_d)^{H\times W}$ as a product manifold, with the particular choice of weight-sharing feature functions $(g_{\theta_k, i, j})_{i, j, k}$ defined by
\begin{equation}
    g_{\theta_k, i, j}: E\in(\mathbb{B}_d)^{H\times W} \mapsto g_{\theta_k} (E_{i,j})\in \mathbb{R}, \quad 
\end{equation}
for every ${1\leq i\leq H}$,  ${1\leq j\leq W}$ and ${1\leq k\leq K}$.

\section{Implementation Details}
\label{sec:supp_implementation}

\paragraph{Segmentation.} Our segmentation model is composed of a DeepLabV3 model, based on a ResNet101 backbone and pretrained on ImageNet, together with the horospherical classification head from \cite{berg2024horospherical}. We train it on 60 epochs with a batch size of 4 and a learning rate of 0.01. It is fed with RGB images of size $512\times 512$, obtained by randomly cropping the original images. We choose a hyperbolic embedding dimension of 2 in all our experiments, and we study the impact of the dimension in Supplementary Materials \ref{sec:supp_dim}. The prototypes from the horospherical head are optimized on the sphere, together with all the other weights, using a Riemannian stochastic gradient descent algorithm \citep{bonnabelStochasticGradientDescent2013}.The weighting term $\beta$ in \cref{eq:loss_segm} is set to $1e-5$. During incremental steps, the new prototypes are initialized according to \cref{eq:new_prototypes}, with $\epsilon_j\sim \mathcal{N}(0,\sigma^2I_d)$ and dispersion $\sigma^2=0.1$.

\paragraph{Change detection.} The change detection part of HyperFlow is trained on the embeddings produced by the segmentation model (which is, for this CD part, frozen) and learns the vector field introduced in \cref{sec:method_busemann_flow}. As we choose to reuse the segmentation prototypes to define this vector field and to fix them, the only thing that needs to be learned is the network $n_\omega$ that we implement with a UNet architecture. The model is trained with an ADAM optimizer, with learning rate $10^{-4}$, batch size 4 on 20 epochs. 

\paragraph{Incremental settings.} As more incremental steps are considered, the segmentation model's training is resumed for 60 more epochs with the same hyperparameters. Note that its architecture is slightly different as its (hyperbolic) head now contains new prototypes. The change detection model's architecture is then redefined to take into account these new prototypes, which leads to a higher number of channels in the UNet, and is retrained from scratch. 

For HiChange, classes are introduced progressively according to the incremental scheme summarized in \cref{tab:tasks}. The table reports only the classes introduced at each step; the full class set at step 
$t$ includes all classes introduced in previous steps together with those added at step $t$. Newly introduced classes are added simultaneously rather than one after another. Class notation is consistent with \cref{fig:hierarchy_labels}. 

For SECOND and Hi-UCD-mini, no incremental setting is used. The tasks $T^0_6$ and $T^0_9$ correspond to the standard setting where all dataset classes are considered from the beginning.

\begin{table}[h!]
\centering
\caption{Incremental tasks for the HiChange dataset. The table summarizes the classes considered at each incremental step.}
\label{tab:tasks}
\begin{tabular}{@{}ccl@{}}
\toprule
\textbf{Step} & \textbf{Task} & \textbf{Classes introduced} \\ \midrule
0             &      $T^0_{4}$         &            $[1,2,3,4]$                 \\
1             &    $T^1_{7}$         &            $[31,32,33,34]$              \\
2             &      $T^2_{13}$         &            $[10,11,12,13,14,15,16]$               \\
3             &       $T^3_{16}$         &            $[40,41,42,43]$               \\
4             &       $T^4_{20}$         &            $[20,21,22,23,24]$         \\
5             &      $T^5_{22}$         &            $[321,322,323]$            \\
6             &     $T^6_{32}$         &            $[110,111,112,113,114,121,122,123,141,142,143]$            \\ \bottomrule
\end{tabular}

\end{table}

\paragraph{Semantic change detection maps and consistency between change and segmentation maps.} Following established work in SCD, our formulation deliberately keeps (and evaluates) the change map and the semantic segmentation maps as separate outputs rather than combining them into a single \emph{from-to} transition map. The latter would couple the errors of individual predictions without providing additional information. The notion of change consistency in our work should therefore be understood at the level of the incremental task rather than as a hard pixel-wise constraint. HyperFlow does not enforce change and semantic prediction to be consistent at every pixel. However, as the semantic taxonomy evolves, both semantic segmentation and change detector evolve to enable the change maps to describe changes within the new semantic classes considered.

\section{Additional Qualitative Results}
\label{sec:supp_qualitative}
This section presents additional qualitative results illustrating the behavior of HyperFlow across datasets. 

\subsection{HiChange}

\begin{figure}
    \centering
    \includegraphics[width=\linewidth]{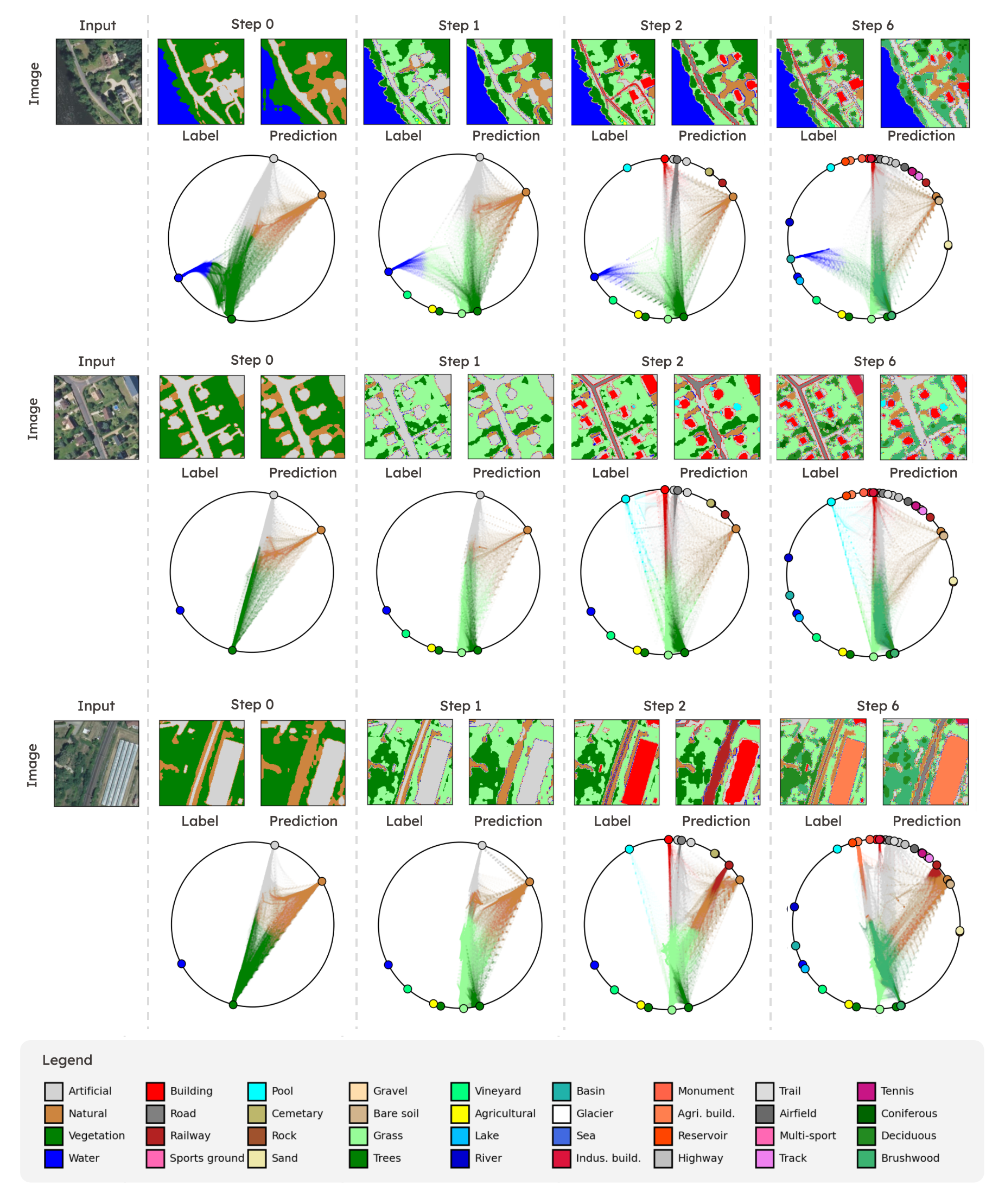}
    \caption{Qualitative examples on HiChange of segmentation across incremental steps and associated embedding space visualizations. For each step, we show the predicted segmentation together with the corresponding hyperbolic embeddings, illustrating how the embedding space evolves as the taxonomy is refined and new classes are introduced. The color of each embedding corresponds to the ground-truth semantic label.}
    \label{fig:supp_incremental}
\end{figure}

For the HiChange dataset, we provide additional visualizations illustrating different aspects of the proposed approach. First, \cref{fig:supp_incremental} presents additional examples of segmentation prediction across incremental steps, together with the associated hyperbolic embeddings. These examples illustrate the capacity of HyperFlow to capture semantic information of increasing complexity as well as the evolution of the organization of the embedding space as the taxonomy is progressively refined. 

\begin{figure}
    \centering
    \includegraphics[width=0.8\linewidth]{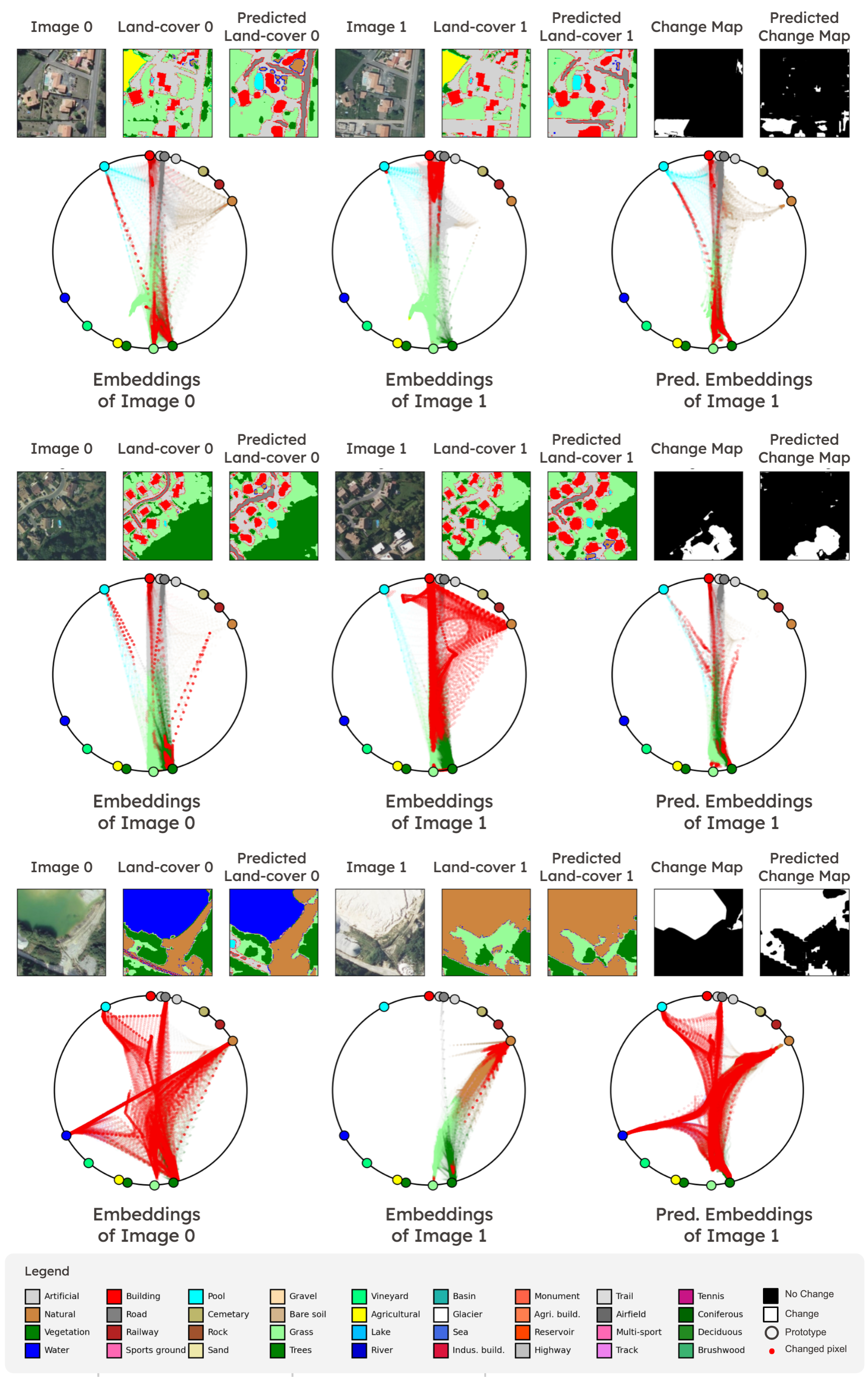}
    \caption{Qualitative examples of SCD at task $T^2_{15}$. For each image pair, we show the predicted segmentation at both timesteps, the predicted change map, the embeddings of the two images and the predicted embeddings of the second image. Pixels corresponding to true semantic changes (in red) exhibit larger displacement in the embedding space, that are not captured by the learned flow and therefore not accounted for in the predicted embeddings.}
    \label{fig:supp_CD}
\end{figure}

Second, \cref{fig:supp_CD} presents additional examples of semantic change detection for the second incremental step ($T^2_{15}$). For each pair of images, we show the predicted segmentation at both timestamps, the resulting change map, and the embeddings of the first and second image, as well as the predicted embeddings of the second image. These examples illustrate how pixels corresponding to changes exhibit larger displacements in the hyperbolic space compared to unchanged regions, enabling HyperFlow to detect changes in an unsupervised way. 

\begin{figure}
    \centering
    \includegraphics[width=\linewidth]{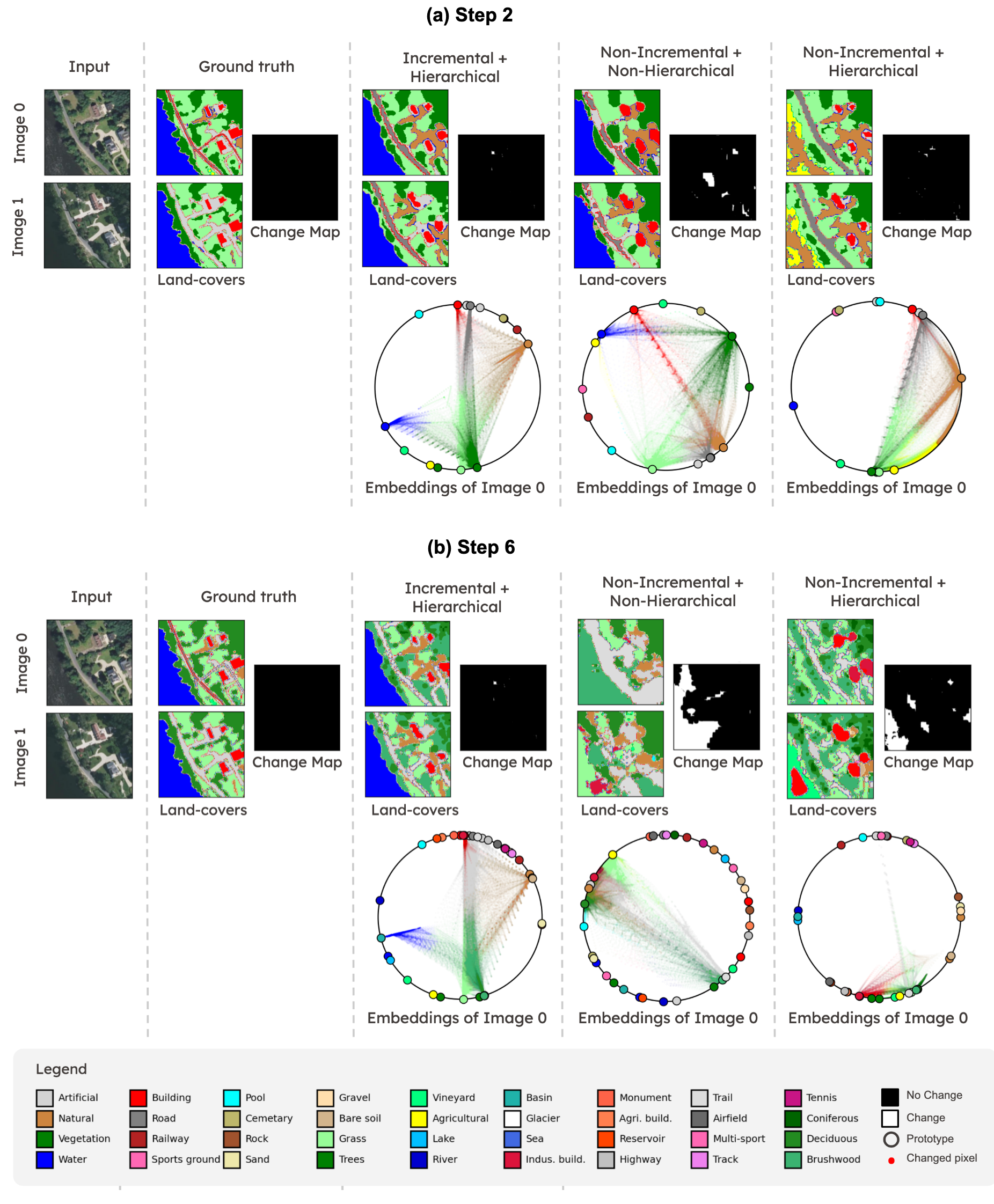}
    \caption{Comparison between incremental and non-incremental scenarios. We visualize the segmentation and change predictions at (a) $T_{13}^2$ and (b) $T_{32}^6$, along with corresponding embedding spaces. Incremental refinement preserves the hierarchical organization, enabling better prediction, whereas directly learning fine-grained categories leads to less structured embeddings and lower performances.}
    \label{fig:supp_incr_vs_non_incr}
\end{figure}

Finally, \cref{fig:supp_incr_vs_non_incr} compares both prediction and embedding spaces obtained under incremental and non-incremental settings. These visualizations provide an intuitive illustration of the effect discussed in the main paper: incremental refinement preserves the global hierarchical organization of the embedding space whereas directly training on the full fine-grained taxonomy tends to produce less structured representations. For the non incremental but hierarchical scenario, the hierarchical organization is also preserved. However, because the model must learn all fine-grained categories from scratch, the task is harder and the resulting performance is lower.

\subsection{SECOND}

\begin{figure}
    \centering
    \includegraphics[width=\linewidth]{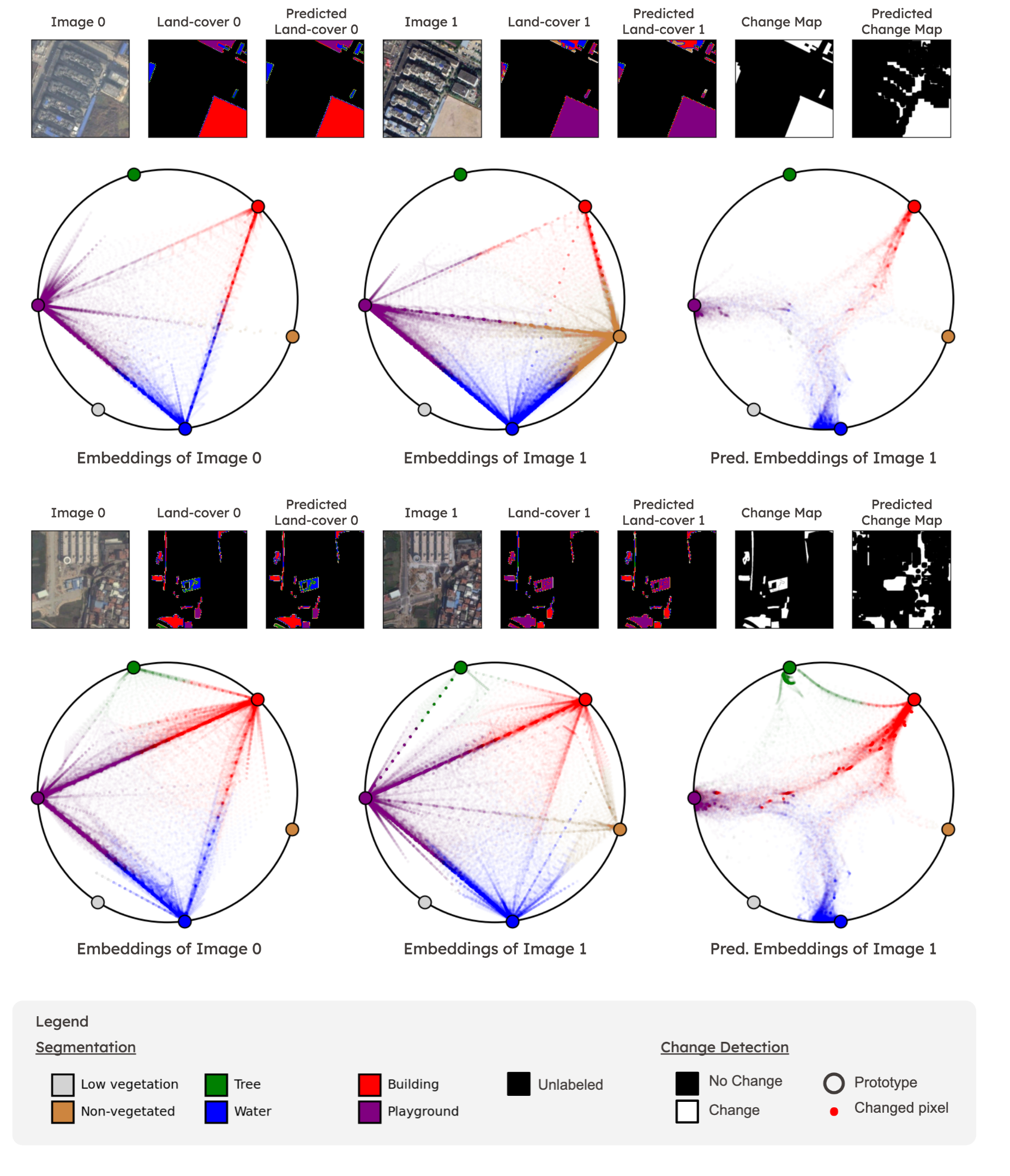}
    \caption{Qualitative examples on the SECOND dataset. For each image pair, we show segmentation and change prediction, and the associated embeddings.}
    \label{fig:supp_visu_SECOND}
\end{figure}

For the SECOND dataset, incremental scenarios are not possible. We therefore present qualitative examples of SCD only. In \cref{fig:supp_visu_SECOND}, we visualize the predicted segmentation at both timesteps, the predicted change map and the embeddings of the first and second image, compared with the predicted embeddings of the second image. 

\subsection{Hi-UCD-mini}

\begin{figure}
    \centering
    \includegraphics[width=\linewidth]{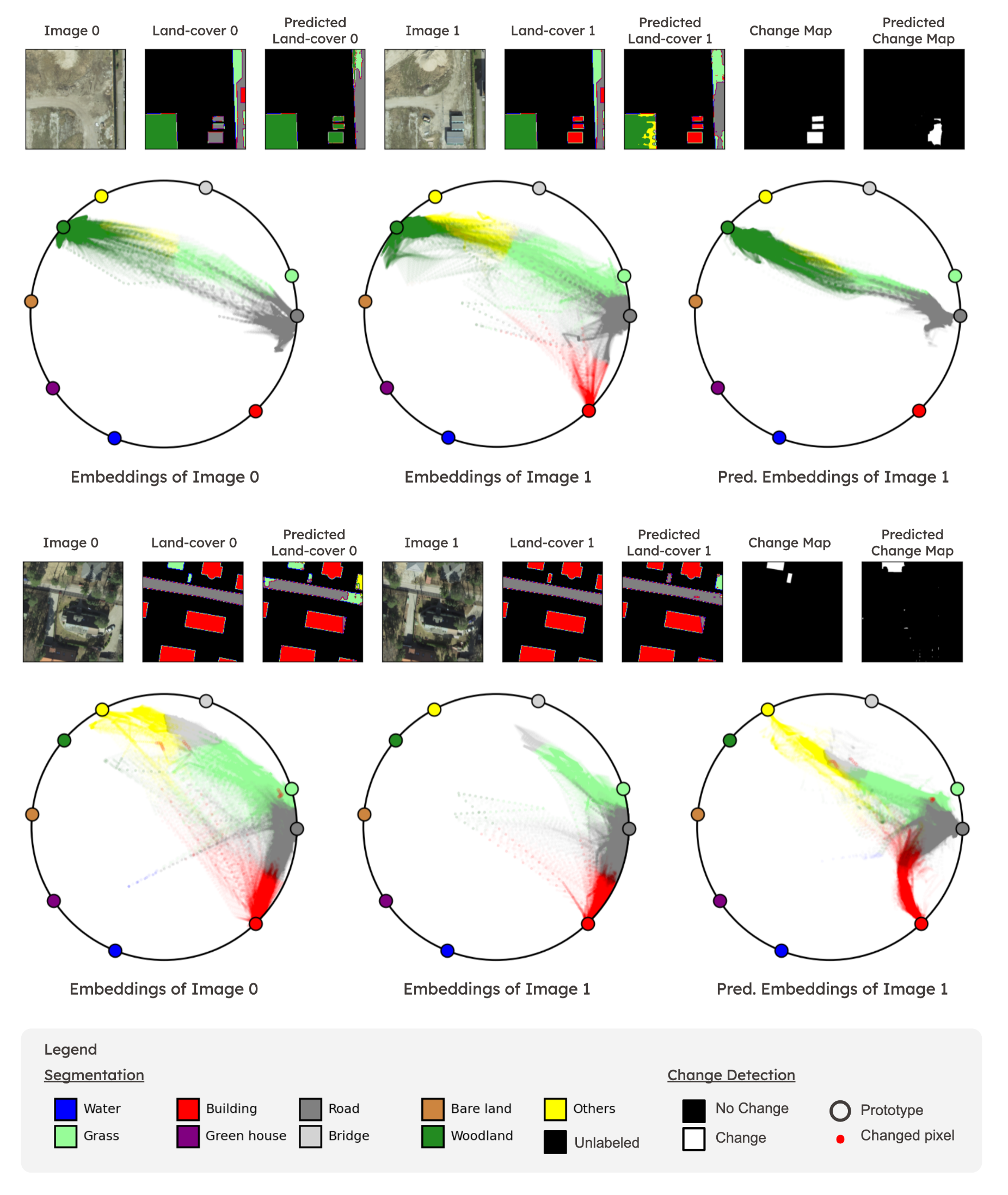}
    \caption{Qualitative examples on the Hi-UCD-mini dataset. For each image pair, we show segmentation and change prediction, and the associated embeddings.}
    \label{fig:supp_visu_hiucd}
\end{figure}

Similarly, for the Hi-UCD-mini dataset, we provide qualitative examples of SCD in \cref{fig:supp_visu_hiucd}. We show for each pair of images the segmentation and change predictions, as well as the associated embedding space visualizations. 

\section{Additional Experiments}
\label{sec:supp_experiments}

\subsection{Influence and Choice of the Threshold}
\label{sec:supp_influence_threshold}
\begin{figure}[!h]
    \centering
    \includegraphics[width=0.9\linewidth]{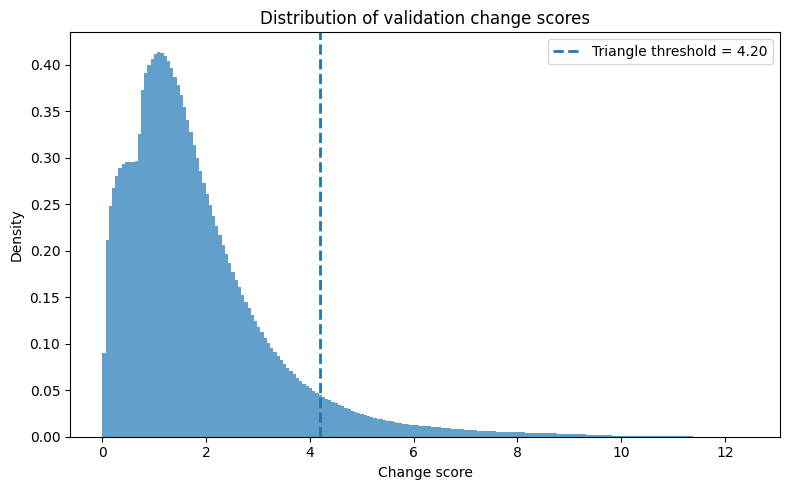}
    \caption{Distribution of change scores and triangle threshold for the HiChange dataset and the base task $T_4^0$.}
    \label{fig:distribution_change_scores}
\end{figure}

The CD part of HyperFlow, based on hyperbolic semantic flows, needs a threshold at inference to transform a continuous reconstruction error into a binary prediction. To obtain binary CD from the distance between the embeddings without relying on ground-truth labels, we require an unsupervised thresholding method. Otsu's method \citep{ostu}, although commonly used, is unsuitable here, as it assumes a bimodal histogram, whereas our distance distribution is dominated by the unchanged mode with a much smaller tail of change scores, as observed in \cref{fig:distribution_change_scores}. Under this imbalance, Otsu does not provide a suitable threshold. We instead use the Triangle algorithm \citep{trianglethreshold}, which is designed for exactly this type of skewed distribution: it selects the threshold as the point of maximum perpendicular distance between the histogram and the line connecting its peak to its tail. 

\begin{figure}[h!]
    \centering
    \includegraphics[width=0.7\linewidth]{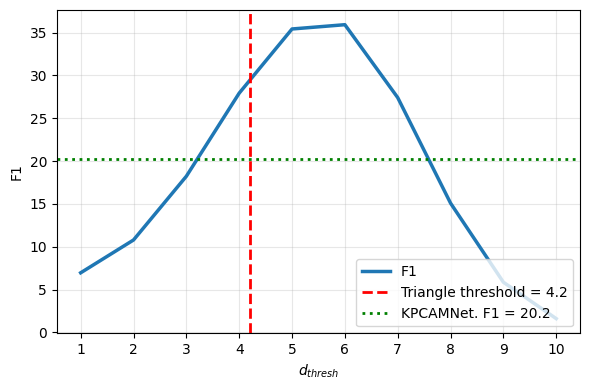}
    \caption{Influence of $d_\mathrm{thresh}$ on CD performance. We present the F1 score of HyperFlow with respect to the threshold used, for HiChange and the base task $T_4^0$. Results are compared with the best unsupervised CD baseline, KPCAMNet.}
    \label{fig:thresh_impact}
\end{figure}

The triangle method, computed on the validation set, for HiChange and the base task $T_4^0$ gives the value $d_\mathrm{thresh}=4.2$. We study the impact of this threshold in \cref{fig:thresh_impact}. $4.2$ appears to be a relevant choice, although not being the best one. While the performance depends on the choice of the threshold, we note that the range in which our model outperforms other unsupervised methods remains large. We therefore keep this value for all experiments involving HiChange. The threshold value is however adapted for the other two datasets, and when using SAM3 and AnySat as embeddings encoder, as the distribution of the change scores may vary significantly in those cases. Similarly, when considering the Euclidean vector fields, the thresholds are adapted to the range of change scores values encountered. The values retained are summarized in \cref{tab:threshold_values}.

\begin{table}[!h]
\centering
\caption{Thresholds values obtained from the triangle method. Unless stated otherwise, these values are used for all experiments.}
\label{tab:threshold_values}
\begin{tabular}{@{}cccc@{}}
\toprule
\textbf{Dataset} & \textbf{Embeddings} & \textbf{Hyperbolic/Euclidean} & \textbf{Threshold} \\ \midrule
HiChange         & Segmentation        & Hyp.                          &   4.2                 \\
HiChange         & Segmentation        & Euc.                          &        0.2            \\
HiChange         & Segmentation        & Euc-Bus.                      &   0.2                 \\
HiChange         & SAM3                & Hyp.                          &  6.3                  \\
HiChange         & AnySat              & Hyp.                          &    3.6                \\
SECOND           & Segmentation        & Hyp.                          &    7.0                \\
SECOND           & SAM3        & Hyp.                          &      6.0              \\
SECOND           & AnySat        & Hyp.                          &        5.4           \\
Hi-UCD-mini       & Segmentation        & Hyp.                          &      5.1              \\
Hi-UCD-mini       & SAM3        & Hyp.                          &      6.0              \\Hi-UCD-mini       & AnySat        & Hyp.                          &          5.6         \\\bottomrule
\end{tabular}
\end{table}

\subsection{Influence of the Backbone Architecture} 
\label{supp:influence_backbone}
All experiments presented in the main paper rely on a single backbone (Deep-LabV3 with a ResNet-101 backbone pre-trained on ImageNet) to ensure fair comparison across methods. We now investigate how the choice of backbone affects both segmentation and change detection performance. In particular, we consider a simpler UNet backbone. Results are provided in \cref{tab:supp_diff_backbones}. As expected, the performance obtained with the UNet backbone is lower than with DeepLabV3. This difference is mainly explained by the smaller model size ($1.3$M parameters compared to $58$M), and the absence of pretraining. Consequently, segmentation performance decreases, which impacts CD and SCD, as good change detection relies on the quality of the underlying segmentation. Nevertheless, it can be observed that for $T_4^0$, the CD performance for the UNet backbone (F1$=14.5$) remains similar to those of the unsupervised methods reported in \cref{tab:CD_perfs}. This result indicates that HyperFlow still provides effective CD even when relying on a weaker segmentation model.

\begin{table}[h!]
\centering
\caption{Influence of the backbone architecture on the performances of HyperFlow.}
\label{tab:supp_diff_backbones}
\begin{tabular}{@{}ccccccccc@{}}
\toprule
\textbf{Metric}                            & \textbf{Backbone}  & $T^0_4$ & $T^1_7$ & $T^2_{13}$ & $T^3_{16}$ & $T^4_{20}$ & $T^5_{22}$ & $T^6_{32}$ \\ \midrule
\multirow{2}{*}{\textbf{Segm. {[}mIoU{]}}} & \textbf{DeepLabV3} & 62.0 & 56.2 & 33.7 & 28.4 & 23.5 & 20.7 &  16.1       \\
                                           & \textbf{UNet}      &   50.5  &  24.3  &   12.3  &     9.1      &  8.0        &   6.9         & 4.7           \\
\multirow{2}{*}{\textbf{CD {[}F1{]}}}      & \textbf{DeepLabV3} & 29.6 & 29.6 & 29.8 & 29.2& 27.7 & 31.6 &  26.0      \\
                                           & \textbf{UNet}      &   14.5  &   16.3  &   17.1     &  17.0         &  17.2         &  18.8         &   17.3        \\
\multirow{2}{*}{\textbf{SCD {[}SCS{]}}}    & \textbf{DeepLabV3} & 39.0 & 34.4 & 22.4 & 19.9 & 17.3 & 17.6 & 13.8        \\
                                           & \textbf{UNet}      &  31.1   &  16.7  &   10.3      &   8.8       &   8.6        &  8.5         &   7.1       \\ \bottomrule
\end{tabular}
\end{table}



\subsection{Do we really need to use the segmentation prototypes?}

\label{sec:appendix_other_prototypes}

The current formulation reuses the Busemann prototypes learned for segmentation as anchors for the vector field. An alternative is to use uniformly distributed prototypes, either learned or fixed. While this provides greater flexibility and allows HyperFlow to operate without segmentation-derived embeddings, which is particularly relevant for non-semantic CD tasks or when using general embeddings not obtained from segmentation pretext tasks, as in \cref{tab:CD_perfs} with SAM3 and AnySat, it may discard part of the semantic information encoded in the learned segmentation prototypes. \Cref{tab:non_segm_proto} compares CD performance using prototypes derived from the segmentation task and prototypes independent of segmentation, both with and without prototype learning. While using non-segmentation prototypes leads to a performance drop, the resulting CD performance remains competitive, demonstrating that HyperFlow can also operate on prototypes not derived from a segmentation pretext task. Learning the prototypes partially closes the gap with segmentation-based prototypes, although not entirely. This can be explained by the fact that CD ultimately concerns changes in semantic content. Segmentation embeddings and prototypes therefore provide a natural basis for modeling this semantic flow.

\begin{table}[h!]
\centering
\caption{Comparison of segmentation-based and independent prototypes for CD on the HiChange dataset, showing the impact of prototype initialization on HyperFlow performance.}
\label{tab:non_segm_proto}
\begin{tabular}{@{}cccccc@{}}
\toprule
\textbf{Segmentation prototypes}    & \textbf{Trained}    & \textbf{Acc.} & \textbf{F1} & \textbf{Pre.} & \textbf{Rec.} \\ \midrule
\ding{51} & \ding{55}   &    92.9           &      29.2       &       20.2        &     52.6          \\
\ding{55} & \ding{55}  &       93.1        &      26.9       &        19.1       &       45.5        \\
\ding{55} & \ding{51}      &      93.9         &      27.9       &       20.8        &     42.5          \\ \bottomrule
\end{tabular}
\end{table}

\subsection{Does working in Hyperbolic space help?}

One of the main hypotheses of this work is that exploiting hyperbolic geometry can improve the CI-SCD task. While \cite{hindel_taxonomy-aware_2024} demonstrated the benefits of hyperbolic representations for CISS on natural images, it remains unclear whether these advantages also extend to change detection (CD) in remote sensing imagery. To investigate this question, we consider two Euclidean counterparts to the proposed Busemann vector field framework. The first preserves the overall methodology while replacing the hyperbolic Busemann function with its Euclidean analogue, namely a scalar product. This variant therefore retains the same formulation and training procedure, differing only in the underlying geometry. The second removes the Busemann formulation entirely, as it is no longer required in Euclidean space, and instead trains a neural network to directly predict the next embeddings. We refer to these two variants as \textit{Euc-Bus.} and \textit{Euc}, respectively, and compare their performance on the CD task with the proposed hyperbolic approach in \cref{tab:hyp_vs_euc}. 

We observe a clear superiority of the hyperbolic formulation. However, in our experiments, the threshold obtained using the triangle-based method for the Euclidean variants was not optimal. This can be attributed to the quite different distribution of embeddings in Euclidean space: compared to hyperbolic embeddings, the embeddings exhibit smaller pairwise distances, as those distances do not grow exponentially. To provide a fairer comparison and remove artifacts of thresholding behavior, we therefore manually selected the optimal threshold for both the \textit{Euc-Bus.} and \textit{Euc} formulations. \Cref{tab:hyp_vs_euc_best_thr} reports their performance using these thresholds, alongside that of the standard hyperbolic HyperFlow with its optimal threshold ($d_\mathrm{thresh}=6$, see \cref{fig:thresh_impact}). Although the performance gap is reduced, the overall trend remains unchanged: the hyperbolic version outperforms its Euclidean counterparts.

\begin{table}[h!]
\centering
\caption{Comparison of hyperbolic and Euclidean formulations for HyperFlow on the HiChange dataset and the base task $T_4^0$, using (a) triangle thresholds and (b) optimal thresholds.}

\begin{subtable}[t]{0.48\textwidth}
    \centering
    \caption{}
    \label{tab:hyp_vs_euc}
    \begin{tabular}{@{}ccccc@{}}
    \toprule
    \textbf{Space}    & \textbf{Acc.} & \textbf{F1} & \textbf{Pre.} & \textbf{Rec.} \\ \midrule
    \textbf{Hyp.}     &    92.9           &      29.6       &       20.5        &     52.8          \\
    \textbf{Euc-Bus.} &      82.1         &    18.9         &           10.8    &       74.7        \\
    \textbf{Euc.}     &     83.1          &      19.7       &         11.4      &       73.4        \\ \bottomrule
    \end{tabular}
\end{subtable}
\hfill
    \begin{subtable}[t]{0.48\textwidth}
        \centering
        \caption{}
    \label{tab:hyp_vs_euc_best_thr}
    \begin{tabular}{@{}ccccc@{}}
    \toprule
    \textbf{Space}    & \textbf{Acc.} & \textbf{F1} & \textbf{Pre.} & \textbf{Rec.} \\ \midrule
    \textbf{Hyp.}     &    97.1           &      35.9       &       47.1        &     29.0          \\
    \textbf{Euc-Bus.} &      94.4         &    34.9         &           25.8    &       53.8       \\
    \textbf{Euc.}     &     94.4          &      35.5       &         26.5      &       53.7        \\ \bottomrule
    \end{tabular}
    \end{subtable}

\end{table}

\begin{table}[]
\centering

\end{table}

\subsection{Embedding Dimension}
\label{sec:supp_dim}
We evaluate the impact of the embedding dimension 
$d$ on performance for the base task $T^0_4$. Results are reported in \cref{tab:supp_dimensions}. Increasing the embedding dimension does not provide a clear improvement on both semantic segmentation and CD performance. The results fluctuate across dimensions, without showing a clear superiority. This behavior is likely influenced by the choice of the threshold that might not be optimal in other dimensions. In our experiments, we used the same threshold value ($d_\mathrm{thresh}=4.2$) for all dimensions, while 
the optimal threshold can be expected to vary slightly with the embedding dimension. 

\begin{table}[h!]
\centering
\caption{Impact of the embedding dimension $d$ on segmentation, CD, and SCD performance for HiChange and the base task $T^0_4$.}
\label{tab:supp_dimensions}
\begin{tabular}{@{}cccc@{}}
\toprule
\textbf{Dimension} & \textbf{Segm. {[}mIoU{]}} &\textbf{ CD {[}F1{]}} & \textbf{SCD {[}SCS{]}} \\ \midrule
\textbf{2}         & 62.0             & 29.6        & 39.0          \\
\textbf{16}        & 60.8             & 21.7        & 37.1          \\
\textbf{64}        & 62.2             & 28.4        & 39.8          \\
\textbf{128}       & 58.6             & 32.6        & 41.8          \\ \bottomrule
\end{tabular}

\end{table}

Despite these results and the fact that higher dimension could potentially provide better  performances, especially if we consider the SCD performance, we keep $d=2$ in the main experiments for three reasons. First, this configuration already surpasses competing methods. Second, $2$-dimensional embeddings enable direct visualization of the learned representations, which provides useful insights into the semantic structure and the learned flow. Third, it reduces the number of parameters in the model. While higher dimensions slightly improve the SCD (SCS) score, the gains remain limited and do not justify the additional parameters.

\subsection{Failure Cases}

\begin{figure}[!h]
    \centering
    \includegraphics[width=\linewidth]{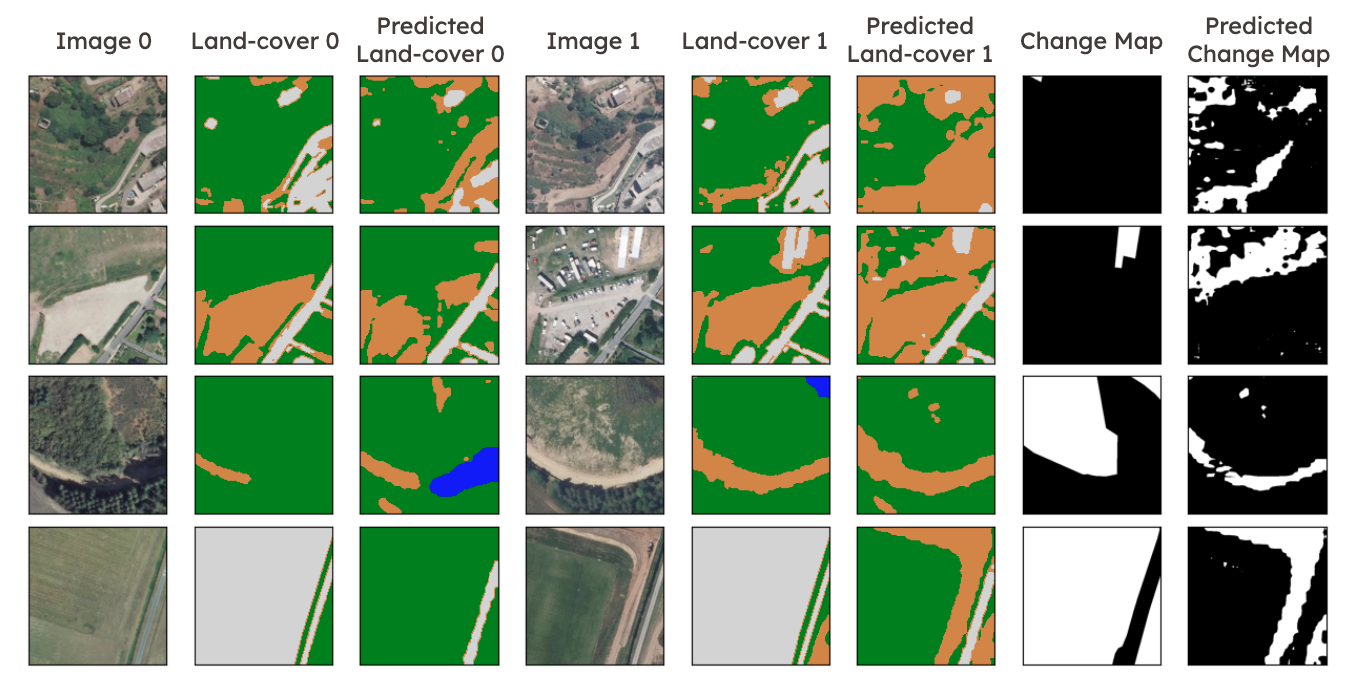}
    \caption{Failure cases of HyperFlow. The first two examples illustrate errors caused by inaccuracies in the predicted land-cover, which propagate to the change map. The last two illustrate cases where the ground-truth annotations present errors.}
    \label{fig:sup_failure_cases}
\end{figure}

Figure \ref{fig:sup_failure_cases} presents typical examples of failure cases. Two main sources of errors can be identified. The first is errors in the predicted segmentation. As illustrated by the first two examples, inaccuracies in the predicted land-cover propagate to the change map. This is consistent with the results discussed in \cref{sec:results_part1}: the quality of the embeddings directly affects the quality of the resulting CD, and, for the segmentation embeddings used here, this quality is closely related to the accuracy of the land-cover segmentation. The second reason comes from errors in the ground-truth annotations, as observed in the last two examples.

\subsection{Computational Complexity}

Our model has $59.2$M parameters ($225.6$ MB) and requires $276.7$ GFLOPs per forward pass.

\end{document}